\documentclass[conference]{IEEEtran}
\IEEEoverridecommandlockouts
\usepackage{cite}
\usepackage{amsmath,amssymb,amsfonts}
\usepackage{algorithmic}
\usepackage{graphicx}
\usepackage{textcomp}
\usepackage[dvipsnames]{xcolor}
\usepackage[hyphens]{url}
\usepackage{microtype}
\usepackage{graphicx}
\usepackage{subfigure}
\usepackage{booktabs} % for professional tables
\usepackage{xspace}
\usepackage{amsmath}
\usepackage{booktabs}
\usepackage{multirow}
\usepackage{hyperref}
\usepackage{enumitem}
\usepackage{comment}
\usepackage{soul}
\usepackage{flushend}

\newcommand{\loka}{LoKA\xspace}
\newcommand{\lokaprobe}{LoKA Probe\xspace}
\newcommand{\lokamodule}{LoKA Mod\xspace}
\newcommand{\lokamodules}{LoKA Mods\xspace}
\newcommand{\lokadispatch}{LoKA Dispatch\xspace}

\newcommand{\code}[1]{\texttt{#1}}
\newcommand{\bluestrikethrough}[1]{}

\newcommand{\revision}[1]{#1}
\newcommand{\shepherd}[1]{#1}

\def\BibTeX{{\rm B\kern-.05em{\sc i\kern-.025em b}\kern-.08em
    T\kern-.1667em\lower.7ex\hbox{E}\kern-.125emX}}
\begin{document}

% Ensure letter paper
\pdfpagewidth=8.5in
\pdfpageheight=11in

%%%%%%%%%%%---SETME-----%%%%%%%%%%%%%
\newcommand{\iscasubmissionnumber}{389}
%%%%%%%%%%%%%%%%%%%%%%%%%%%%%%%%%%%%

\pagenumbering{arabic}

%%%%%%%%%%%---SETME-----%%%%%%%%%%%%%
\title{\loka: Low-precision Kernel Applications for Recommendation Models At Scale}
%\author{\normalsize{ISCA 2026 Submission
%    \textbf{\#\iscasubmissionnumber} -- Confidential Draft -- Do NOT Distribute!!}}
%%%%%%%%%%%%%%%%%%%%%%%%%%%%%%%%%%%%

\author{
    \IEEEauthorblockN{
        Liang Luo, Yinbin Ma, Quanyu Zhu, Vasiliy Kuznetsov, Yuxin Chen, Neng Shi, \\ 
        Jian Jiao, Jiecao Yu, Buyun Zhang, Tongyi Tang, Xiaohan Wei, Yanli Zhao, Zeliang Chen, \\ 
        Yuchen Hao, Venkatesh Ranganathan, Sandeep Parab, Yantao Yao, Maxim Naumov, \\
        Chunzhi Yang, Shen Li, Ellie Wen, Wenlin Chen, Santanu Kolay, Chunqiang Tang
    }
    \vspace{0.5em} % Adds a little breathing room before the affiliation
    \IEEEauthorblockA{
        \textit{Meta AI} \\
        Menlo Park, CA, USA \\
        \{liangluo, yinbin, qyz, vasiliy, yuxinc, jianj, jiecaoyu, buyunz, tongyitang, ubimeteor, \\
        yanlizhao, zlc, haoyc, verangan, sandeepparab, yantao, mnaumov, zorror, \\
        shenli, ellie.wen, wenlinchen, skolay, tang\}@meta.com
    }
}

\maketitle
\thispagestyle{plain}
\pagestyle{plain}

%%%%%% -- PAPER CONTENT STARTS-- %%%%%%%%

\begin{abstract}

Recent GPU generations deliver significantly higher FLOPs using lower-precision arithmetic, such as FP8. While\bluestrikethrough{ FP8 has been} successfully applied to large language models (LLMs), its adoption in large recommendation models (LRMs) has been limited. This is because LRMs are numerically sensitive, dominated by small matrix multiplications (GEMMs) followed by normalization, and trained in communication-intensive environments. Applying FP8 directly to LRMs often degrades model quality and prolongs training time. These challenges are inherent to LRM workloads and cannot be resolved merely by introducing better FP8 kernels. Instead, a system-model co-design approach is needed to successfully integrate FP8.
We present \loka (Low-precision Kernel Applications), \shepherd{a framework that makes FP8 practical for LRMs through three principles: profile under realistic distributions to know where low precision is safe, co-design model components with hardware to expand where it is safe, and orchestrate across kernel libraries to maximize the gains. Concretely,} \lokaprobe is a statistically grounded, online benchmarking method that learns activation and weight statistics, and quantifies per-layer errors. This process pinpoints safe and unsafe, fast and slow sites for FP8 adoption. \lokamodules is a set of reusable model adaptations that improve both numerical stability and execution efficiency with FP8. \lokadispatch is a runtime that leverages the statistical insights from \lokaprobe to select the fastest FP8 kernel that satisfies the accuracy requirements. Deployed on production LRMs that serve billions of users with advertising recommendations at a major social media company, \loka delivers up to 20\% higher training throughput and 40\% faster inference on heterogeneous GPUs (H100, B200, \shepherd{GB200}, MI300X and \shepherd{MI350X}) in production environment, with no quality loss, turning FP8 from a model-quality risk into a reliable performance lever for LRMs at scale.

\end{abstract}

\section{Introduction}
\label{label:intro}

Lower-precision numerics, such as FP8 and FP4, have been major drivers of GPU performance improvements due to their efficient use of chip area and power. As shown in Table~\ref{tab:nvidia_tensor_performance}, the B200's TF32 FLOPs is 7× that of the A100, while its FP8 and FP4 FLOPs are 29× and 58× the A100's TF32 FLOPs, respectively. To fully leverage the high FLOPs of lower-precision numerics, ML models must adopt them effectively without sacrificing quality.

\begin{table}[h]
\footnotesize
\centering
\resizebox{0.7\columnwidth}{!}{%
\begin{tabular}{lccc}
\toprule
TFLOPs & \textbf{TF32} & \textbf{BF16} & \textbf{FP8/INT8} \\
\midrule
A100~\cite{nvda_a100}      & 156   & 312   & 624    \\
H100 NVL~\cite{nvda_h100}  & 418   & 840   & 1,671    \\
B200 HGX~\cite{nvda_b200}  & 1,100 & 2,250 & 4,500 \\
%\midrule
%\multicolumn{5}{l}{\textbf{Same-Precision Performance Scaling}} \\
%\multicolumn{5}{c}{7$\times$ cumulative scaling with 2--3$\times$ generational improvements} \\
%\addlinespace
%\multicolumn{5}{l}{\textbf{Mixed-Precision Performance Scaling}} \\
%\multicolumn{5}{c}{58$\times$ cumulative scaling enabled by progressively lower precision arithmetic} \\
%\multicolumn{5}{c}{with improvements 2--3$\times$ with each new precision data type} \\
\bottomrule
\end{tabular}%
}
\caption{NVIDIA GPU Dense Peak Performance.}
\label{tab:nvidia_tensor_performance}
\vspace{-1em}
\end{table}

While \revision{recent work~\cite{deepseekai2025deepseekv3technicalreport,hernandez2025towards}} demonstrated that FP8 can be applied effectively to large language models (LLMs), to our knowledge, no prior work has achieved successful FP8 training for large recommendation models (LRMs) at scale. LRMs are critical industrial workloads that power applications such as advertising, short-video, and news-feed ranking. They predate LLMs and differ substantially in architecture and numerical behavior. Applying FP8 to LRMs is particularly challenging: on a state-of-the-art framework, TorchAO~\cite{or2025torchaopytorchnativetrainingtoservingmodel}, we observe 1.3× slowdown and up to 2.5\% relative log loss degradation (note that 0.02\% relative log loss is considered significant in production) when applying FP8 training to a representative production recommendation model, due to the following characteristics.

\begin{figure}[!t]
    \centering
    \includegraphics[width=0.8\linewidth]{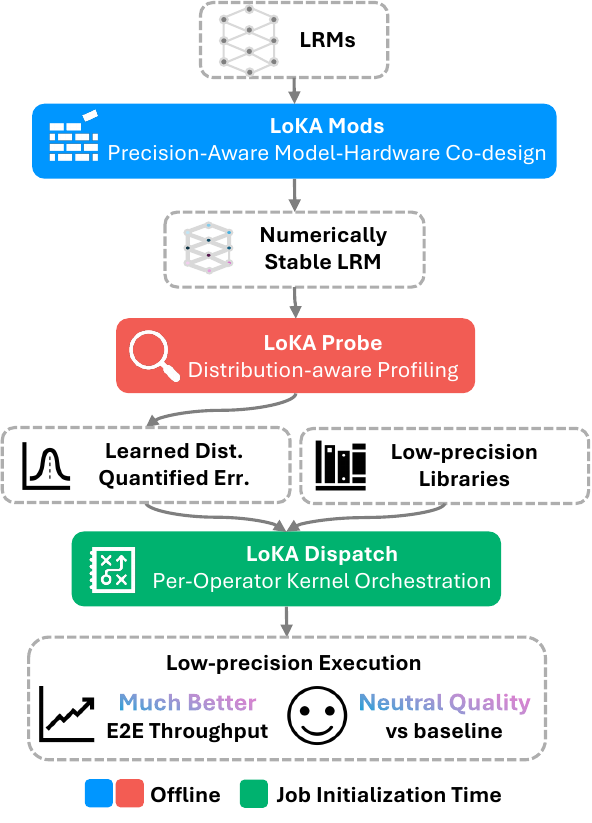}
    \caption{\shepherd{\loka Overview}}
    \label{fig:overview}
    \vspace{-2em}
\end{figure}

First, LRMs operate under tight quality constraints—even minor degradations in model accuracy are unacceptable, leaving little room for approximation.
Second, LRMs are architecturally heterogeneous. Unlike LLMs’ relatively uniform Transformer~\cite{vaswani2017attention} stacks, LRMs combine wide ensembles~\cite{10.1145/3580305.3599846}, deep hierarchical stacking~\cite{zhang2022dhendeephierarchicalensemble,zhang2024wukongscalinglawlargescale}, and specialized interaction modules~\cite{10.1145/3580305.3599846}, each exhibiting distinct numerical sensitivities.
Third, LRMs exhibit low arithmetic intensity despite LLM-scale FLOPs complexity~\cite{pmlr-v235-zhai24a}. They consist primarily of many small GEMMs followed immediately by normalization layers, where quantization overhead can dominate and often negate the benefits of low-precision execution. \bluestrikethrough{Fourth, LRM training heavily favors data parallelism~\cite{interformer,zhang2024wukongscalinglawlargescale,dmt,neo} due to their heterogeneity and their much smaller effective batch size per GEMM compared to LLMs, with the latter employing usually O(10M) tokens per update~\cite{deepseekai2025deepseekv3technicalreport}.}

These challenges cannot be solved merely by introducing better low-precision kernels, as the fundamental problems are inherent to how LRMs are structured and trained. What is required is a system-model co-design approach that can diagnose where low precision is viable, modify model components to remain stable under reduced precision, and select the right kernels at runtime to maximize efficiency \shepherd{and maintain} quality.

We present \loka (Low-precision Kernel Applications), a framework designed to unlock the benefits of FP8 and emerging precisions for large-scale recommendation models \shepherd{}.

\shepherd{\loka is built on top of three principles (Figure~\ref{fig:overview})}:

\begin{enumerate}[leftmargin=*, labelindent=0pt, labelsep=0.5em, align=parleft]
\item \shepherd{\textbf{Distribution-Aware Profiling — Know where low precision is safe}. Standard library benchmarks use synthetic inputs and overestimate low-precision readiness. Real workload distributions, especially confounded by complex LRMs, can be heavy-tailed, correlated, or non-stationary. These expose quantization errors that random tensors miss. Any low-precision adoption should begin by profiling under realistic distributions to identify per-operator risk. To that end, \loka introduces \textit{\lokaprobe}, a statistically grounded benchmarking method that identifies which layers in an LRM can safely operate at low precision efficiently.}

\item \shepherd{\textbf{Precision-Aware Model-Hardware Co-design — Expand where low precision is safe}. When profiling reveals that certain operators are numerically unsafe or inefficient under low precision, the solution is not to fall back to high precision universally but to co-design model components with hardware execution constraints. This means strategically modifying operations so that they are simultaneously more numerically stable and more hardware-efficient. We propose \textit{\lokamodules}, a collection of redesigned modules that improve numerical stability under low-precision execution, reducing the risk of divergence or quality degradation.}

\item \shepherd{\textbf{Per-Operator Kernel Orchestration — Maximize the gains from low precision}.  No single low-precision library or recipe dominates across all operator shapes and hardware. Treating each operator as an independent optimization problem — selecting the fastest implementation that satisfies an accuracy constraint — consistently outperforms any uniform policy. \loka adopts \textit{\lokadispatch}, a unified runtime mechanism that integrates multiple low-precision libraries and dynamically selects the fastest kernel that satisfies both accuracy and throughput constraints.}
\end{enumerate}

% Together, these components transform low-precision arithmetic from a brittle optimization into a dependable performance lever for production-scale LRMs.
\shepherd{Together, these principles ensure \loka remains effective across a wide range of hardware and beyond specific workloads. To demonstrate this,} we leverage actual training data, comprising thousands of features, along with realistic batch sizes and production-scale model configurations and assess \loka across three state-of-the-art recommendation model families integrating a wide range of industrial and open-source architectures ~\cite{zhang2022dhendeephierarchicalensemble, vaswani2017attention, Zhang_2024, wang2021dcn, dcpp, dlrm}: Wukong~\cite{zhang2024wukongscalinglawlargescale}, Interformer~\cite{interformer} and External Large FM (ELFM)~\cite{liang2025externallargefoundationmodel}. Across diverse scales and hardware generations including Nvidia H100, B200 and GB200 as well as AMD MI 300X \revision{and 350X} clusters, \loka consistently delivered significant end-to-end gains. It enabled up to 20\% faster training throughput and 40\% faster inference throughput in our hyperscale production environment, all while preserving model quality. \revision{\loka was successfully launched to flagship models, serving billions of users at a major social media company for both training and inference.} These results demonstrate that low-precision computation, when applied through a systematic framework, can serve as a practical and reliable accelerator for production-scale recommendation systems.
\section{Low Precision Computation for Recommendation Models}
\label{sec:background}
This section examines the unique challenges of applying low-precision training to recommendation models, highlighting key differences from language models that make direct adoption of existing techniques ineffective.

\subsection{Large Recommendation Models}

\begin{figure}[t]
    \begin{center}
    \includegraphics[width=0.8\linewidth]{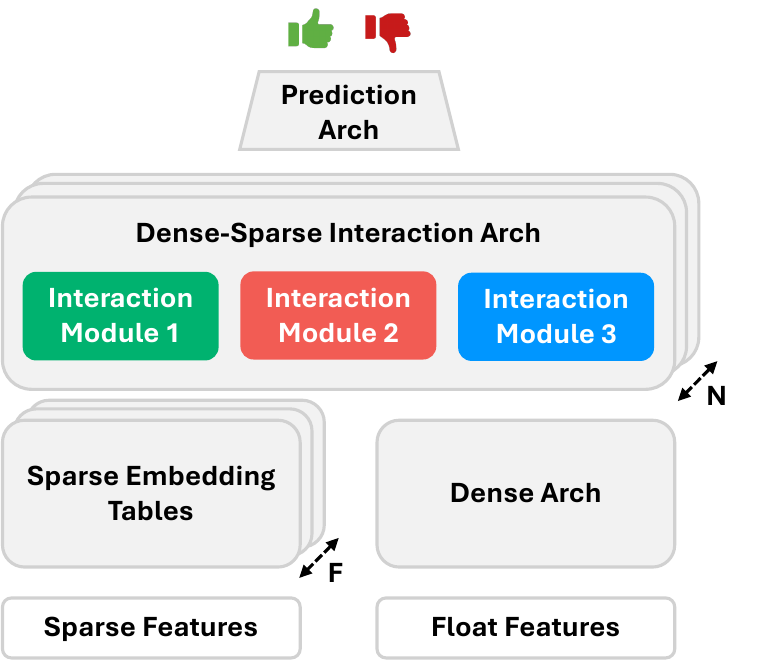}
    \caption{A typical model architecture of a LRM.}
    \label{fig:lrm}        
    \end{center}
    \vspace{-1em}
\end{figure}

Modern recommendation models, illustrated in Figure~\ref{fig:lrm}, process diverse inputs to predict user-item interactions. They comprise four main components:

\begin{itemize}[leftmargin=*, labelindent=0pt, labelsep=0.5em, align=parleft]
    \item \textbf{Sparse embedding tables} store dense vectors for discrete features like user IDs and item categories. These tables contain most model parameters and are distributed across GPUs using techniques like embedding sharding.
    \item \textbf{Dense architecture} processes continuous numerical features through standard neural network layers.
    \item \textbf{Dense-sparse interaction modules} combine embeddings and dense features through stacked layers containing specialized interaction components including deep cross networks~\cite{wang2021dcn}, multi-head attention~\cite{afn}, factorization machines~\cite{zhang2024wukongscalinglawlargescale}, or ensembles~\cite{zhang2022dhendeephierarchicalensemble} thereof. They dominate computational complexity in the LRM.
    \item \textbf{Prediction layers} generate final scores from features.

\end{itemize}

Like LLMs, LRMs follow scaling laws~\cite{zhang2024wukongscalinglawlargescale,kaplan2020scaling,ardalani2022understanding,geng2023vip5,guo2023embedding} and have reached comparable FLOPS complexity. However, unlike LLMs that process relatively static data requiring single training runs, LRMs must continuously adapt to evolving user preferences and new items through online training~\cite{gao2025deck}, making both training and inference performance critical.

Modern LRMs use hybrid training~\cite{neo}: sparse embedding tables are sharded across GPUs~\cite{neuroshard}, and are communicated via the AlltoAll collectives~\cite{neo, dmt}, while dense components use Fully Sharded Data Parallel (FSDP)~\cite{fsdp}.

\subsection{Low-precision Training and Inference}

\begin{figure}[t]
  \centering
  % Top: Figure
  \begin{minipage}{\columnwidth}
    \centering
    \includegraphics[width=0.65\columnwidth]{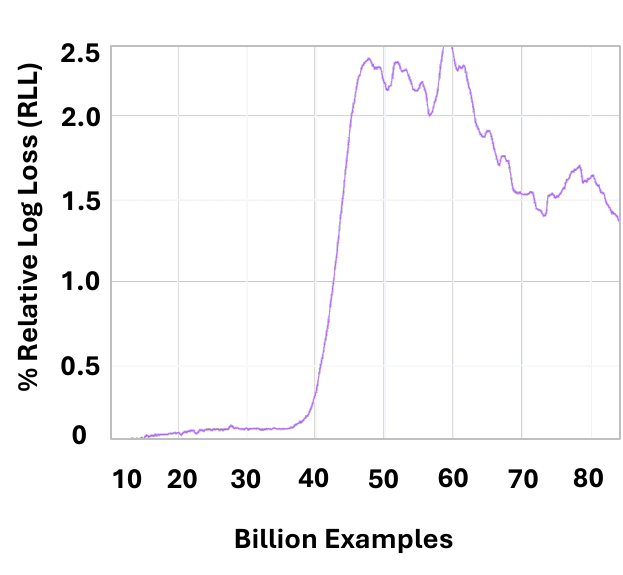}
  \end{minipage}
  % Bottom: Table
\begin{minipage}{\columnwidth}
  \centering
  \resizebox{\columnwidth}{!}{%
    \begin{tabular}{cccc}
      \toprule
      \textbf{BF16 Throughput} & \textbf{FP8 Throughput} & \textbf{FP8 Slowdown} & {\textbf{Relative Log Loss (\%)}} \\
      \midrule
      520K/s & 395K/s & 1.32$\times$ & 2.5 \\
      \bottomrule
    \end{tabular}%
  }
\end{minipage}
  \caption{Significant Relative Log Loss (top) and throughput (bottom) degradation are observed when training a Wukong model in FP8 low precision.}
  \label{fig:og_rll}
  \vspace{-1em}
\end{figure}

Low-precision approaches span several categories~\cite{gholami2021surveyquantizationmethodsefficient}. Quantization-Aware Training (QAT) simulates quantization effects during training while maintaining full-precision weights, improving accuracy but not training speed. Post-Training Quantization (PTQ) applies quantization to trained models without additional training, offering simplicity but typically larger accuracy loss. Post-Quantization Training (PQT) combines PTQ with fine-tuning to recover accuracy. Direct Low Precision (DLP) employs native low-precision training and inference, offering the greatest throughput benefits. DLP is the focus of \loka, though it presents the greatest technical challenges because it requires native low-precision throughout entire training and inference.

Despite successful FP8 adoption in LLMs, direct application to LRMs fails dramatically. To demonstrate this, we applied TorchAO~\cite{or2025torchaopytorchnativetrainingtoservingmodel}, a widely-used PyTorch low-precision library to all linear layers (input/output features $\geq$ 128) in the state-of-the-art Wukong model using 64 H100 GPUs with tensorwise FP8 scaling.

Figure~\ref{fig:og_rll} shows the results: FP8 training incurs 1.3× slowdown and up to 2.5\% relative log loss degradation. This failure stems from fundamental differences between LRMs and LLMs highlighted in \S\ref{label:intro}.

\begin{table}[h]
\footnotesize
\centering
\resizebox{0.8\columnwidth}{!}{%
\begin{tabular}{lcc}
\toprule
             & \textbf{Normal Dist.} & \textbf{LRM Input Dist.} \\
\midrule
Pytorch BF16 & 0.03 & 0.04 \\
TorchAO RW   & 0.47 & 0.53 \\
DeepGEMM BW  & 0.49 & 0.56 \\
FBGEMM RW    & 0.48 & 0.52 \\
\bottomrule
\end{tabular}%
}
\vspace{1em}
\caption{Geometric mean of MERE of 4 important linear layers in a LRM compared to TF32 baseline for low-precision kernels with different input distributions.}
\label{tab:rll_bf16_fbgemm_rll}
\vspace{-1em}
\end{table}

\subsection{Characterizing Low-Precision Quantization Overheads and Error in LRMs}
\label{sec:error_characterization}

Standard low-precision libraries optimize for accuracy and performance using random \revision{or fixed-distribution~\cite{githubDeepGEMMteststest_bf16pyMain,githubAotestkerneltest_blockwise_tritonpyMain,githubFBGEMMfbgemm_gputestquantizefused_8bit_rowwise_testpyMain}} tensor benchmarks and common GEMM shapes~\cite{or2025torchaopytorchnativetrainingtoservingmodel}. However, \revision{recent study~\cite{dettmers2022llmint88bitmatrixmultiplication} has demonstrated that at scale, models develop highly systematic outlier activations that severely disrupt standard quantization techniques and} real LRM workloads deviate substantially from these test conditions.

We analyzed a four-layer production LRM with linear layers, comparing three state-of-the-art low-precision libraries—FBGEMM~\cite{khudia2021fbgemmenablinghighperformancelowprecision}, TorchAO, and DeepGEMM~\cite{deepgemm}. For each library, we evaluated two input distributions:
(1) standard normal (as used in typical benchmarks), and
(2) learned distributions obtained from actual LRM training traces.

Using the learned distributions allows us to synthesize representative inputs and quantify how errors manifest on out-of-distribution data, the conditions that are rarely exercised by standard benchmarks. We enabled fast accumulation, sampled 50 input–weight pairs per layer, and computed average statistics to ensure significance. Only forward performance was tested for simplicity.

\noindent\textbf{Error Analysis}. For each linear layer, we measure the FP8 execution error relative to a TF32 reference using Mean Element-wise Relative Error (MERE): \[ \text{MERE}(\text{out}^{(M,N)}, \text{ref}^{(M,N)}) = \sum_{m}^M\sum_{n}^N |\frac{\text{out}_{m,n}-\text{ref}_{m,n}}{\text{ref}_{m,n}}| \]

MERE captures the average element-wise deviation of low-precision outputs from full-precision results.

Findings. Table~\ref{tab:rll_bf16_fbgemm_rll} summarizes results: the geometric mean of MERE across layers increases by up to 15\% when tested with realistic LRM distributions versus standard normal inputs. This demonstrates that out-of-distribution activations significantly amplify quantization errors, precisely the failure mode not revealed by conventional random benchmarks.

\begin{figure}[t]
    \begin{center}
    \includegraphics[width=0.9\linewidth]{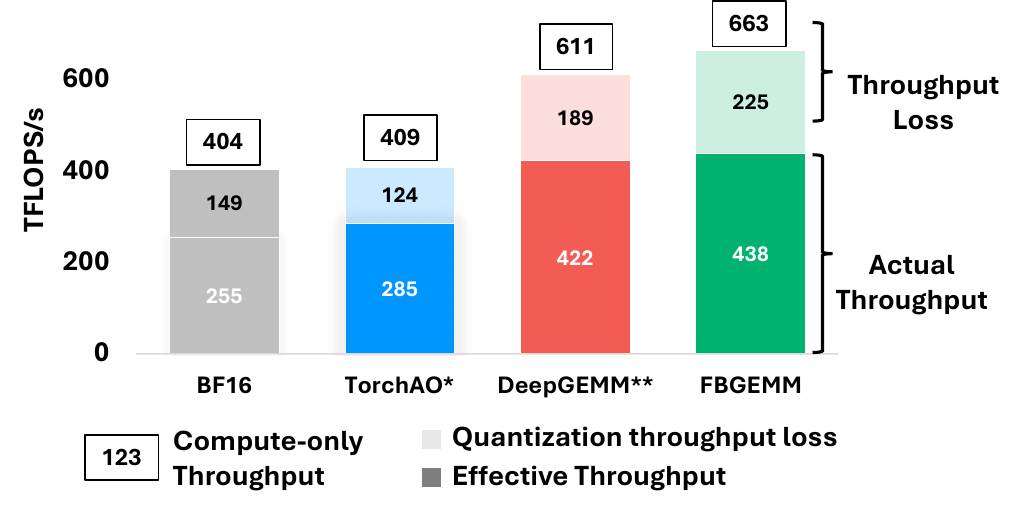}
    \caption{Compute throughput ablation of low-precision kernels on representative LRM shapes. \revision{*/**: Padding overheads not included in both TorchAO and DeepGEMM; torch.compile not available for DeepGEMM.}}
    \label{fig:fp8_overheads}        
    \end{center}
    \vspace{-1.5em}
\end{figure}

\noindent \textbf{Throughput Analysis} We benchmarked 27 GEMM shapes from production LRMs, ranging from (2048, 256) @ (256, 768)\revision{\footnote{Denote a matmul of two tensors with shapes (2048, 256) and (256, 768)}.} to (2048, 123200) @ (123200, 1024) on H100 GPUs. Figure~\ref{fig:fp8_overheads} shows results separated into pure compute throughput versus end-to-end performance including quantization overhead. Key findings:

\begin{itemize}[leftmargin=*, labelindent=0pt, labelsep=0.5em, align=parleft]
    \item End-to-end FP8 speedup over BF16 is limited to 1.6× on average.
    \item Maximum effective TFLOPS/s remains under 20\% of hardware capacity.
    \item Quantization overhead consumes over 30\% of end-to-end GEMM latency.
    \item Including memory allocation overhead \revision{(e.g., layout manipulation)}, FP8 can perform worse than BF16.

\end{itemize}

These results demonstrate that applying low-precision computation to LRMs requires addressing both accuracy degradation from realistic data distributions and performance overhead from quantization operations which are challenges that cannot be solved by library improvements alone.

\subsection{The Deployment Status Quo of Low Precision Kernels}

\begin{table}[h]
\footnotesize
\centering
\resizebox{0.55\columnwidth}{!}{%
\begin{tabular}{lccc}
\toprule
             & \textbf{Training} & \textbf{Inference} \\
\midrule
TF32 & 95\% & 0\% \\
BF16/TF32 & 5\% & - \\
FP16      & - & 99\% \\ 
FP8  & 0 & 1\% (PTQ) \\
\bottomrule
\end{tabular}%
}
\caption{Proportion of models trained and served using a specific datatype. - means this choice is not applicable.}
\label{tab:dtypestats}
\vspace{-1em}
\end{table}

Real-world adoption of low-precision training and inference for LRMs remains limited. In a survey of top 500 Ads ranking LRMs at a large social-media company (Table~\ref{tab:dtypestats}), we observed: in training, 95\% of models run in high precision (TF32), 5\% use mixed precision (BF16/TF32), and 0\% train in FP8; in inference, 99\% of models serve in FP16, with only 1\% using FP8 via PTQ. These figures underscore the practical hurdles: numerical stability, small-GEMM quantization overheads, and communication-dominated runtimes, and all these impede FP8 deployment at scale.
\section{\loka}
\shepherd{
As established in \S\ref{label:intro}, the fundamental barriers to low-precision LRM training cannot be addressed by better kernels alone. They require a systematic approach guided by three principles: profiling under realistic distributions to know where low precision is safe, co-designing model components with hardware to expand where it is safe, and orchestrating across kernel libraries to maximize the resulting gains. \loka instantiates each principle as a concrete component. \lokaprobe Probe implements distribution-aware profiling (Principle 1). \lokamodules realizes precision-aware model-hardware co-design (Principle 2). \lokadispatch provides per-operator kernel orchestration (Principle 3). We describe each below.
}
  
% The fundamental issues in low precision training of LRMs stem from how they are structured and trained. Thus, we need a system-model co-design approach that can: (1) diagnose where low-precision is safe and efficient, (2) modify model components for numerical stability under reduced precision, and (3) select optimal kernels at runtime to maximize efficiency without sacrificing quality.

% We present \loka, a framework specifically designed to unlock low-precision benefits for large-scale recommendation models. LoKA consists of three complementary components.

\shepherd{\subsection{\lokaprobe: Distribution-Aware Profiling}}

\begin{figure}[t]
    \begin{center}
    \includegraphics[width=0.8\linewidth]{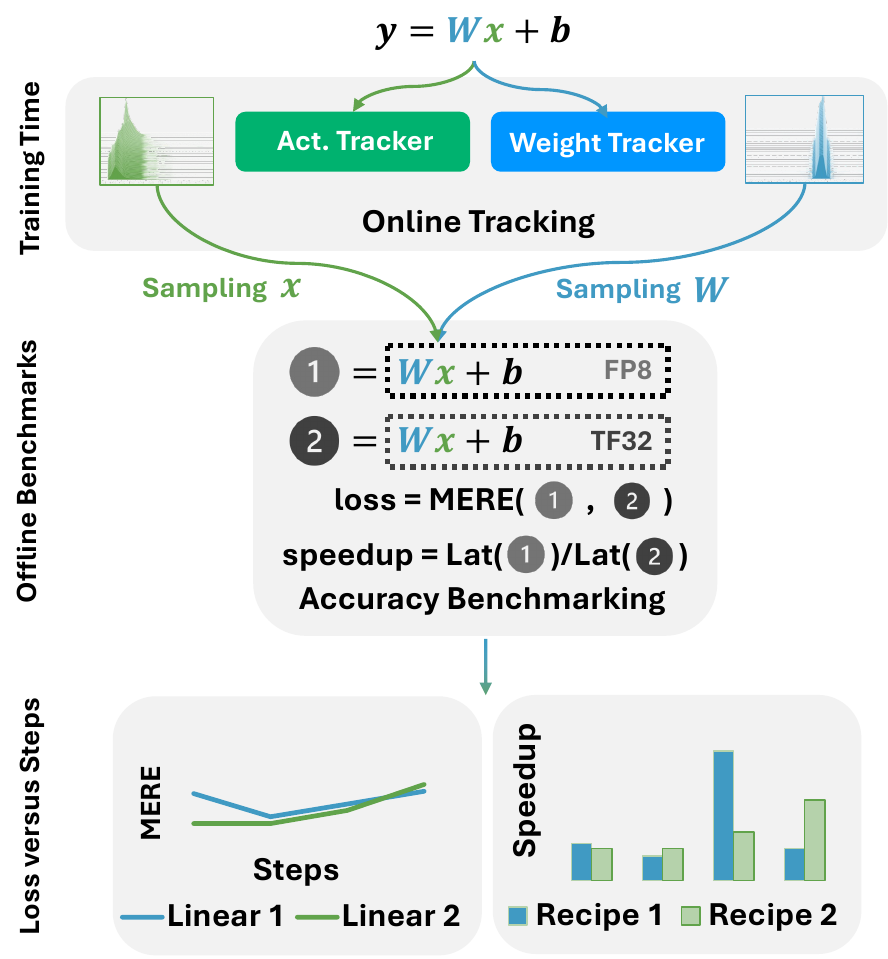}
    \caption{\lokaprobe learns and stores necessary parameters online for offline accuracy benchmarks to derive statistically significant loss for each linear layer under low precision execution.}
    \label{fig:loka_probe}
    \vspace{-1.5em}    
    \end{center}
\end{figure}
The critical insight behind \loka is that \textit{standard low-precision testing fundamentally underestimates real-world quantization errors}. Existing libraries benchmark accuracy using random tensors (typically normal distributions) which fail to capture the complex statistical properties of actual LRM activations and weights. This leads to overly optimistic error estimates and explains why ``battle-tested" low-precision kernels still cause training divergence in practice.

\lokaprobe addresses this by learning the true distributions of inputs and weights during training, then using these learned distributions for statistically significant offline accuracy (and throughput) assessment. This approach reveals quantization errors that random testing misses, enabling informed decisions about which layers can safely use low precision.

\subsubsection{Online Distribution Learning}
\lokaprobe operates in two phases: online learning during training and offline benchmarking for error analysis.

During training, \lokaprobe efficiently tracks the statistical properties of each linear layer's inputs and weights without storing the actual tensors (which would be prohibitively expensive and prone to overfitting). Instead, it maintains compact statistical summaries that can recreate realistic distributions. \lokaprobe models distributions as multivariate Gaussians. For a 2D tensor T of shape (M, N), \lokaprobe samples from \[T \sim \mathbb{G}(\mu, \Sigma)\] where $\mu \in \mathbb{R}^{M \times N}$ is the mean and $\Sigma \in \mathbb{R}^{MN \times MN} $ is the covariance matrix. 

This Gaussian assumption is motivated by both theoretical and empirical observations: activations in large models, after layer normalization or RMS normalization, tend to concentrate
around zero and exhibit light-tailed, symmetric statistics,
which are well captured by a normal distribution.
Moreover, under the central-limit effect, the pre-activation of each neuron aggregates
many independent (or weakly correlated) input contributions, making its distribution approximately Gaussian even in nonlinear regimes.

Finally, LoKA’s objective is not to perfectly reproduce the long-tail statistics of activations but to approximate their \emph{second-order structure} sufficiently for quantization-error analysis. Within this scope, Gaussian families provide analytically closed moments, stable parameter updates, and simple sampling rules,
making them the pragmatic choice for large-scale, online distribution
tracking.

However, while conceptually simple, storing the full covariance matrix ($M^2N^2$ elements) is still intractable for large tensors. Therefore, we seek to significantly reduce \lokaprobe's storage requirements.

\noindent \textbf{Optimized Input Distribution Modeling} Activations play a critical role in selecting layers to quantize~\cite{MLSYS2024_42a452cb}. For them, we exploit the independence of the batch dimension, a key property in recommendation models where cross-batch operators like BatchNorm are avoided to prevent information leakage\footnote{Consider the case when the current (user, item) pair and the user's future iteration (user, item in the future) both present in the same local batch, a batch norm would effectively use the user's \textit{future} behavior, which may already encode their current action in its input feature, to predict their current behavior, leading to catastrophic overfitting.}. By treating the batch dimension as independent, it enables us to model only along the feature dimension, reducing storage from $O(M^2N^2)$ to manageable $O(N^2)$.

Tracking the mean is trivially done in a streaming fashion. For variance, we implement a batched Welford tracker~\cite{chan1983algorithms} that efficiently updates covariance online using only $O(N^2)$ memory, by decomposing the final covariance as follows.

Let the current batch be $X\in\mathbb{R}^{B\times K}$ (rows are samples, columns are features, therefore $B=M$ and $K=N$ using previous annotations).
Define the batch mean $\mu_b\in\mathbb{R}^{K}$ and batch \emph{scatter}:
\[
S_b \;=\; (X-\mathbf{1}_B\mu_b^T)^T\,(X-\mathbf{1}_B\mu_b^T)\;\in\;\mathbb{R}^{K\times K}.
\]
Given historical summaries $(n_{\text{old}},\,\mu_{\text{old}},\,\Sigma_{\text{old}})$, where
$\Sigma_{\text{old}}$ is the (unnormalized) scatter matrix, the merged quantities are
\begin{gather*}
n_{\text{new}} = n_{\text{old}} + B, \\
\delta = \mu_b - \mu_{\text{old}}, \\
\mu_{\text{new}} = \mu_{\text{old}} + \frac{B}{n_{\text{new}}}\,\delta, \\
\Sigma_{\text{new}} = \Sigma_{\text{old}} + S_b + 
\frac{n_{\text{old}}\,B}{n_{\text{new}}}\delta\delta^T.
\end{gather*}
The the unbiased (sample) covariance is
\[
\Sigma \;=\; \frac{\Sigma_{\text{new}}}{n_{\text{new}}-1}\;\;(\text{for } n_{\text{new}}>1).
\]

\noindent\textit{Implementation notes.} \lokaprobe accumulates $\Sigma$ in higher precision (e.g., FP32). %  We then set $\Sigma=\widehat{\mathrm{Cov}}$ for sampling later.

\noindent \textbf{Optimized Weight Distribution Modeling}
We cannot assume dimension independence for $W$ (weight matrix), so the trick for input modeling does not apply. Instead, we model a weight matrix $W\in\mathbb{R}^{M\times N}$ with a matrix-normal distribution:
\[
W \sim \mathcal{MN}\!\big(M,\; U,\; V\big),
\quad \text{and} \quad \mathrm{vec}(W)\sim \mathcal{N}\!\big(\mathrm{vec}(M),\, V\otimes U\big),
\]
where $U\in\mathbb{R}^{M\times M}$ is the row covariance and $V\in\mathbb{R}^{N\times N}$ is the
column covariance. This reduces storage from $O(M^2N^2)$ to $O(M^2+N^2)$
while capturing essential correlations.

Let $W_c = W - M$ be the mean-centered weight. We maintain $(U,V)$ online using a
Kronecker-factor (flip–flop–style) update with exponential moving average (EMA).
To avoid explicit matrix inverses, we use linear solves with (regularized) Cholesky factors.

\emph{Per-update (single minibatch) estimates:}
\begin{align*}
&\text{Solve}\quad L_V L_V^T = V + \varepsilon I,\qquad
\widetilde{W} = W_c\, L_V^{-T}, \\
&U' \;=\; \frac{1}{N}\,\widetilde{W}\,\widetilde{W}^T,
\qquad\qquad\qquad\;\;\;(\text{$M\times M$}) \\[4pt]
&\text{Solve}\quad L_U L_U^T = U + \varepsilon I,\qquad
\widehat{W} = L_U^{-1}\, W_c, \\
&V' \;=\; \frac{1}{M}\,\widehat{W}^T\,\widehat{W},
\qquad\qquad\qquad\;\;\;(\text{$N\times N$})
\end{align*}

\emph{EMA smoothing:}
\begin{gather*}
U'' = m\,U + (1-m)\,U', \\
V'' = m\,V + (1-m)\,V', \\
U \leftarrow \tfrac{1}{2}\!(U'' + U''^T) + \varepsilon I, \\
V \leftarrow \tfrac{1}{2}\!(V'' + V''^T) + \varepsilon I.
\end{gather*}

\emph{Scale identifiability.}
Because $V\otimes U$ is invariant under $(U,V)\mapsto (cU,\, V/c)$ (where $\otimes$ is the Kronecker product~\cite{harville1998matrix}, we renormalize to prevent
drift using $s$ for better numerical stability:
\[
s \;=\; \frac{\mathrm{trace}(U)}{M},\qquad
U \leftarrow \frac{U}{s},\qquad
V \leftarrow s\,V.
\]

These updates allow us to avoid forming $V^{-1}$ or $U^{-1}$ explicitly.
In practice we use small $\varepsilon$ (e.g., $10^{-6}\!\times\!\tfrac{\mathrm{trace}(U)}{M}$) and a momentum
$m\in[0.9,0.99]$ for stable online tracking.

\begin{comment}
we use the matrix normal distribution~\cite{gupta2018matrix} $\mathbb{MN}(\mu, U \otimes V)$ with row covariance $U \in R^{M \times M}$ and column covariance $V \in R^{N \times N}$ to model W's distribution. $\otimes$ is the Kronecker product~\cite{harville1998matrix} in the formula. This reduces storage from O($M^2N^2$) to O($M^2 + N^2$) while capturing essential correlations. We update these using a Kronecker factor-inspired approach with momentum:

\[ U'= W_{\text{centered}} V^{-1} W_{\text{centered}}^T \]
\[ V' = W_{\text{centered}}^T U^{-1} W_{\text{centered}} \]
\[ U'' = mU + (1 - m)U'\]
\[ V'' = mV + (1 - V)V'\]

where similarly, $W_\text{centered}$ is the current mean centered weight matrix. $U', V'$ are the row/column covariance matrix updated using the other covariance, with $U'', V''$ being the final result after the exponential moving average (EMA), and $m$ is the momentum.
\end{comment}

To minimize overhead, \lokaprobe activates every 100 training iterations and asynchronously saves statistical parameters every 10,000 iterations, translating to a negligible ($\leq 1\%$) throughput overhead. This provides comprehensive coverage of distribution evolution throughout training while maintaining minimal performance impact.

\noindent\textbf{Sampling learned distributions.}
Given the tracked statistics $(\mu, \Sigma)$ for activations and $(M, U, V)$ for weights,
\lokaprobe can synthesize representative inputs and weights for offline benchmarking.

\emph{(a) Input sampling.}
We draw a synthetic activation batch $T'\in\mathbb{R}^{B\times K}$ by sampling
\[
Z \sim \mathcal{N}(0, I_K),
\qquad
T' = \mathbf{1}_B\mu^T + Z\,L_\Sigma^T,
\quad L_\Sigma L_\Sigma^T = \Sigma + \varepsilon I,
\]
where $L_\Sigma$ is the Cholesky factor of the (regularized) covariance.%
\footnote{We add a small jitter $\varepsilon I$ (e.g., $10^{-6}\!\times\!\tfrac{\mathrm{trace}(\Sigma)}{K}$)
to maintain numerical stability.}

\emph{(b) Weight sampling.}
For weights modeled as $W\sim\mathcal{MN}(M, U, V)$, we sample
\begin{gather*}
Z \sim \mathcal{N}(0, I_{M\times N}), \quad W' = M + L_U\,Z\,L_V^T, \qquad \\
L_U L_U^T = U + \varepsilon I,\;
L_V L_V^T = V + \varepsilon I.
\end{gather*}
This procedure preserves both row-wise and column-wise second-order correlations,
yielding realistic surrogate weights and activations that match the
training-time statistics without storing full tensors.

\begin{comment}
    \noindent \textbf{Sampling Learned Distributions}
To sample a new input $T'$ and a new weight matrix $W'$ from the distribution for offline benchmarking, \lokaprobe does the following:

\[ T' = \Sigma^{\frac{1}{2}}Z + \mu, Z \sim \mathbb{N}(\textbf{0}, \textbf{1})  \]

\[W' = U^\frac{1}{2}ZV^\frac{1}{2} + \mu, Z \sim \mathbb{N}(\textbf{0}, \textbf{1})\]
\end{comment}

\subsubsection{Offline Error Quantification}
Using the learned distributions, \lokaprobe generates statistically representative test cases and compares low-precision kernel outputs and performance against high-precision references. This reveals layer-specific vulnerabilities that standard random testing misses, and helps us prune layers that do not benefit from low precision acceleration. 

For each linear layer, we compute MERE and speedup using inputs and weights sampled from learned distributions rather than synthetic ones. Layers with high MERE scores or low speedups are flagged as problematic for low-precision execution. Figure~\ref{fig:loka_probe} summarizes this workflow.

% To further reduce overheads during training, \lokaprobe is activated every 100 iterations, and the required parameters for offline, in-distribution sampling, $\Sigma, \mu$ for inputs, and $U, V, \mu$ for weight sampling are stored to a remote persistent storage every 10000 iterations in an asynchronous manner. \lokaprobe keeps track of all historical trend of these parameters to understand long-term fidelity of low-precision computation accuracy during the offline benchmark process.

% \lokaprobe then evaluates each linear layer under low-precision version against their full-precision version, using inputs and weights sampled from the learned input distribution that corresponds to a particular training step. A MERE is then derived for each layer. The linear layers with the highest MEREs are collected for detailed analysis.

\subsubsection{\lokaprobe Key Findings}
We use \lokaprobe to analyze the Wukong model used in \S\ref{sec:error_characterization}, sampling 100 realistic input-weight pairs for each linear layer and identified three critical patterns that result in large losses and unoptimal performance:

\begin{figure}[t]
    \begin{center}
    \includegraphics[width=0.9\linewidth]{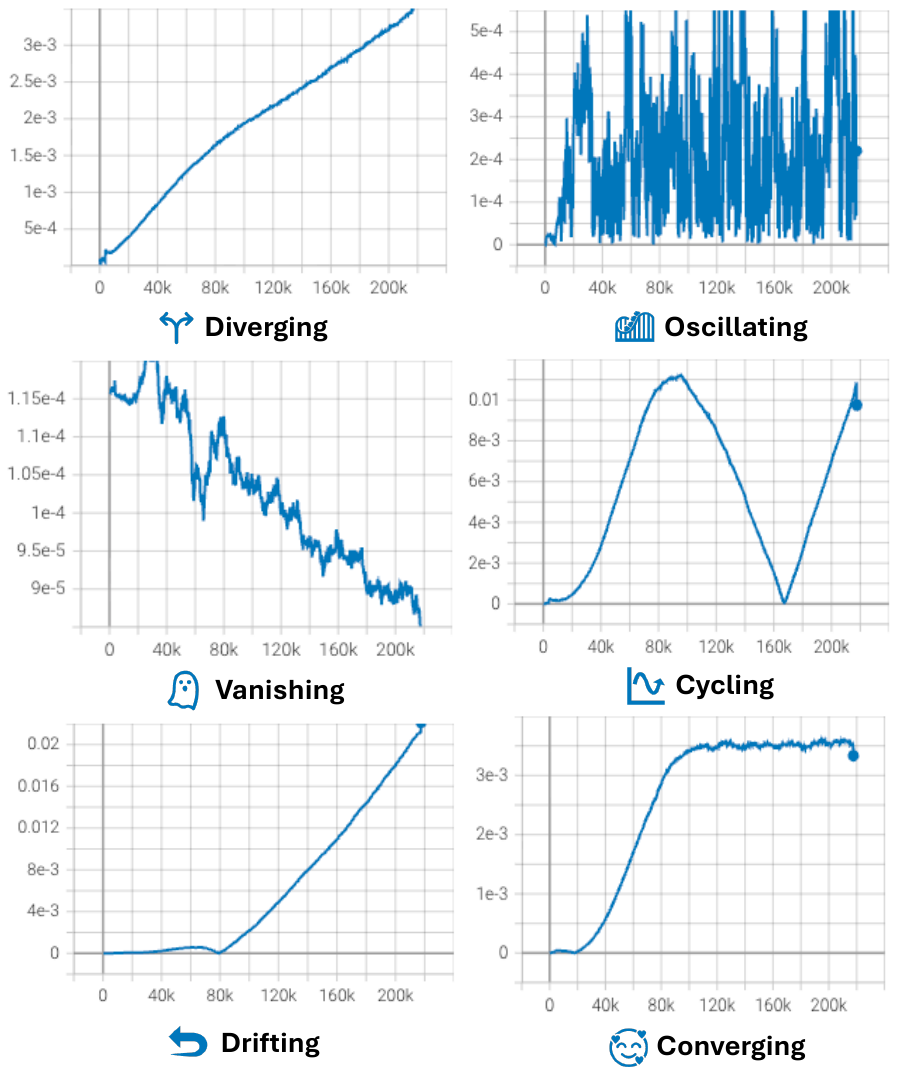}
    \caption{Typical behaviors of bias norm of in Wukong training. Biases can introduce instability during training, especially under low-precision conditions.}
    \label{fig:bias_term_divergence}
    \vspace{-1.5em}        
    \end{center}
\end{figure}

\begin{itemize}[leftmargin=*, labelindent=0pt, labelsep=0.5em, align=parleft]
    \item \textbf{Problematic bias terms} Figure~\ref{fig:bias_term_divergence} shows different modes of L2 bias norm evolution during training. A significant portion of biases never converge, with some reaching values $\geq 0.1$. These diverging biases cascade through subsequent modules, causing out-of-bounds errors. When clamped and quantized, they can cause smaller values to vanish entirely.
    
    % in linear layers. Figure~\ref{fig:bias_term_divergence} shows different modes of how the L2 bias norm evolve with respect to training steps. We observe a significant portion of biases never converge, and some can even go up to significant values ($\geq$0.1), creating significant challenge as it can cascade down to the later modules, leading to out of bound of errors, and, when clamped and quantized later, this can cause smaller values to vanish.

    \item \textbf{Normalization overhead and errors} LayerNorm~\cite{ba2016layernormalization} operations require complex variance computations prone to mean cancellation when values are similar. This forces normalization to run in higher precision, requiring expensive dequantization before LayerNorm and requantization afterward, which is an overhead that often exceeds low-precision benefits, and prevents low-precision execution of such layers. Modern LRMs employs LayerNorms aggressively in between layers in MLP for training stability, creating a sizable impact on the end-to-end latency.

%    Normalization induces additional slowdowns and quantization errors. Wukong employs layer normalization (LayerNorm) ~\cite{ba2016layernormalization} as activation layers in between MLP layers to mean-center inputs to improve low-precision quantization fidelity. Unfortunately, LayerNorm involves complex operations involving summing of squared difference to compute variance, and is prone to mean cancellation if values are large and close together. These force layer noramlization to be performed in higher precision. As a result, we must dequantize before a LayerNorm and quantize again to low precision after it.
    
    \item \textbf{Sigmoid-based activation instability} LRM's heavy use of sigmoid functions (Swish activation\cite{Ramachandran2017SwishAS}: $x\cdot \sigma(x)$, SwishNorm: $x\cdot \sigma(\text{Norm}(x))$) involves exponential operations that amplify large elements while diminishing small ones, dramatically increasing quantization loss.    
\end{itemize}

\shepherd{\subsection{\lokamodules: Precision-Aware Model-Hardware Co-design}}

The vulnerabilities identified by \lokaprobe (problematic bias terms, normalization overhead, and sigmoid instability) cannot be fixed by tightening kernel precision alone. Instead, we take a model-kernel co-design approach, introducing \lokamodules: redesigned building blocks that improve both numerical stability and execution efficiency under low-precision conditions.

% Each LoKA Mod directly addresses a specific vulnerability pattern.

% The traditional approach to fixing the MERE involves tightening the numeric of the kernel. In \loka, we take an alternative, model-kernel co-design approach to tackle these challenges. We introduce a series of low-precision friendly modules, dubbed \lokamodules, to reduce both the execution overheads of the modules and to improve their accuracy under low-precision conditions.

\subsubsection{No Bias}
We draw inspiration from recent LLM architectures that have moved away from bias terms. Models like DeepSeek eliminate biases from all feedforward and normalization layers, while PaLM~\cite{anil2023palm2technicalreport} and Falcon~\cite{almazrouei2023falconseriesopenlanguage} remove them from feedforward layers while retaining them in normalization components. Following this trend, we remove all bias terms from Wukong modules except the final prediction layers, where bias terms can be beneficial for different prediction tasks. \revision{This modification also provides the benefit of potentially reduced communication overheads: with FSDP per-parameter padding~\cite{liang2024torchtitanonestoppytorchnative}, bias tensors smaller than the world size can incur significant overheads.}

\begin{figure}[t]
    \begin{center}
    \includegraphics[width=0.85\linewidth]{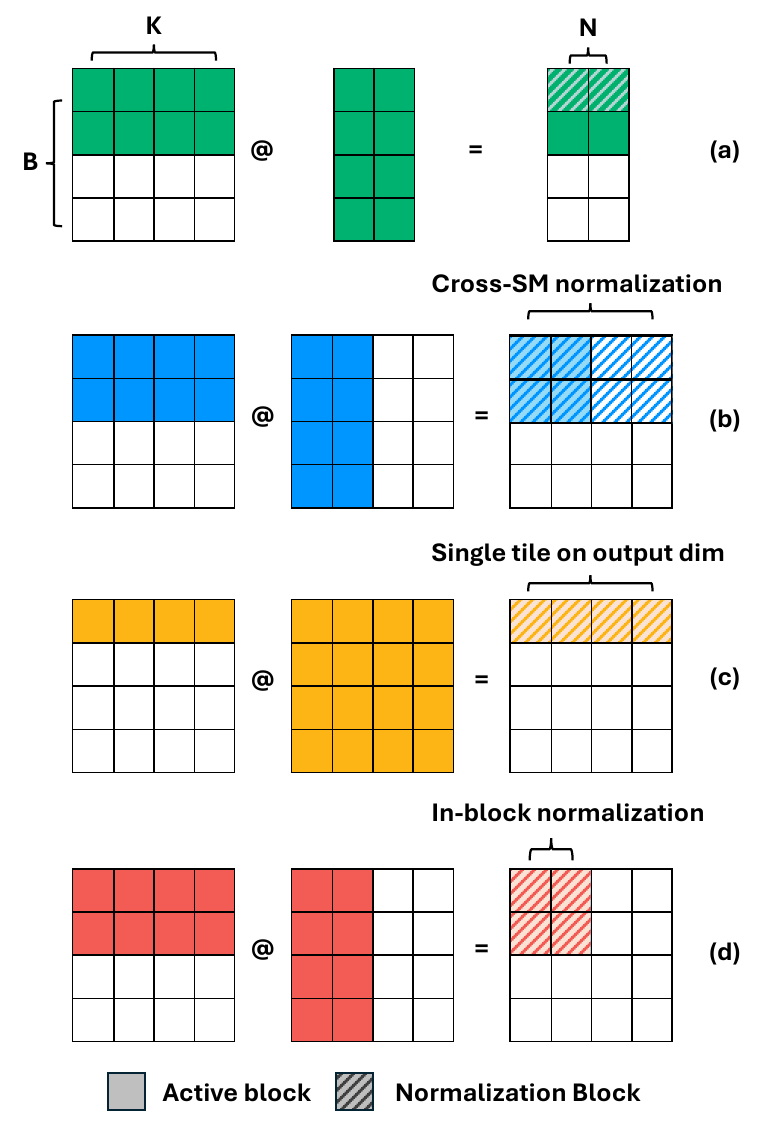}
    \caption{BlockNorm design}
    \label{fig:blocknorm}
    \vspace{-1.5em}        
    \end{center}
\end{figure}

\subsubsection{Block-wise Normalization}
Our objective is to fuse normalization directly into the GEMM epilogue to minimize HBM I/O. \revision{While similar to epilogue fusion~\cite{10.1145/3315508.3329973}, our application is in a different context.} By performing normalization immediately after GEMM completion while the output tiles still reside in on-chip memory (L1/L2 caches or registers), we avoid costly global memory traffic. However, conventional normalization layers operate along the feature dimension, which often misaligns with the physical data layout of GEMM outputs. 

To address this, we \shepherd{proposes} \textbf{BlockNorm}, a normalization approach \shepherd{inspired by GroupNorm~\cite{wu2018groupnormalization} which is originally designed to apply normalization within individual channels of a feature map}. \shepherd{In our version of BlockNorm, instead of computing statistics across the full output dimension or individual feature, it applies \emph{root-mean-square} normalization~\cite{zhang2019rootmeansquarelayer} across predefined blocks along the feature dimension, which happens within each computational block of the GEMM kernel:}
\[
\code{RMSNorm}\big((Wx + b).\code{view}(-1, \text{BlockN})\big).\text{view}(B, N).
\]
We adopt RMSNorm over LayerNorm because it only normalizes by the L2-norm of activations, avoiding mean subtraction and thus reducing catastrophic cancellation errors under low-precision execution.

\noindent\textbf{Case 1: Large batch, small output dimension.}  
When the batch size $B$ is large and the output dimension $N$ is small (Figure~\ref{fig:blocknorm}a), the entire feature vector required for normalization fits within a single thread block. In this regime, BlockNorm behaves identically to standard normalization layers, as all statistics are computed locally. Additional operations such as activation and (de)quantization can also be fused together before the final output is written to HBM.

\noindent\textbf{Case 2: Small batch, large output dimension.}  
When $N$ is large, however, a single thread block can no longer hold an entire output row in shared memory. Computing RMS statistics now requires cross-block synchronization (Figure~\ref{fig:blocknorm}b), which negates most of the performance gains of fusion. One possible mitigation is to reduce the tile size along the batch dimension, but this incurs \emph{SM wave quantization effects}~\cite{smwaveeffect} and lowers the L2 cache hit rate for the $W$ matrix, yielding negligible end-to-end speedup (Figure~\ref{fig:blocknorm}c).

\noindent\textbf{Tradeoff and design choice.}  
We further explored semi-fused approaches where GEMM outputs partial statistics for later normalization, or tiling assignments aligned with the normalization axis at the SM level. While these approaches offer limited gains, they require extensive manual tuning for each new shape, which reduces generality. To achieve robust performance across diverse tensor shapes, \loka deliberately relaxes full mathematical equivalence in favor of efficiency, leading to the final \emph{BlockNorm} formulation (Figure~\ref{fig:blocknorm}d).

Concretely, BlockNorm normalizes over fixed-size blocks (e.g., 256 elements) rather than the full output dimension. During both training and inference, the same block size is used, and normalization statistics are computed independently per block \shepherd{to ensure train-test consistency}. % Activation and (de)quantization then proceed as in the standard fused epilogue. A special case arises when \loka is used with DeepGEMM, where hierarchical accumulation periodically uses higher-precision data types to improve numerical stability. In such cases, the output may require requantization before subsequent operators; although this introduces an additional casting step, it also improves normalization precision by exploiting the temporary high-precision accumulator.

% \begin{enumerate}[leftmargin=*, labelindent=0pt, labelsep=0.5em, align=parleft]

% \item Block-wise GEMM: Tiling is performed along both the batch and output dimension. The output block size is a predefined value (e.g. 256) and consistently used throughout training and inference.

% \item Block-wise Normalization and Activation: Normalization statistics are computed independently within each block. Normalization and activation are then applied using these block-local statistics, followed by activation functions.

% \item Quantization and Dequantization: when block-wise GEMM is tiled along $K$ dimension, \sota kernels such as DeepGEMM perform low-precision multiplication and cast result to Tensor Core higher precision representation (e.g., fast accumulation), and further accumulation along the $K$ dimension is promoted to Cuda Core for higher precision (e.g., FP32). This means the final output needs to be quantized again if the next operator accepts low precision inputs.
% \end{enumerate}

% The difference between standard RMSNorm and our proposed BlockNorm is shown in Figure~\ref{fig:blocknorm} (b) and Figure~\ref{fig:blocknorm} (d).

\begin{figure}[t]
    \begin{center}
    \includegraphics[width=\linewidth]{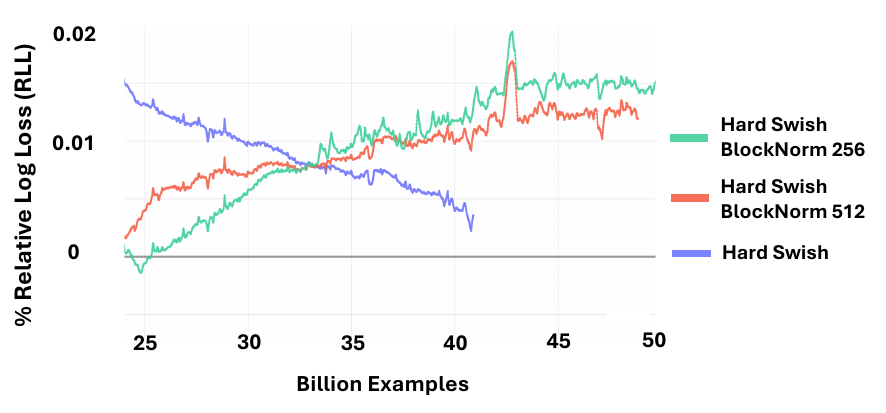}
    \caption{\revision{Hard Swish and BlockNorm with sufficiently large block size converges to minimal loss.}}
    \label{fig:blocknorm256}        
    \end{center}
    \vspace{-2em}
\end{figure}

\noindent \textbf{Striking a Better Numerical and Performance Tradeoff with Model Co-design}
While the actual computation differs between BlockNorm and standard normalization practices, we argue that it 
%, we argue that BlockNorm retains the key properties of RMSNorm that contributes to training/inference stability, including (1) output invariance to input/weight scaling; and (2) gradient invariance to input scaling and negative \revision{correlation with to weight scaling}. This allows BlockNorm to reduce sensitivity to input scaling and implicit adaptation of learning rate to large weight scales.
\revision{does not break RMSNorm: it is mathematically equivalent to an unparameterized Grouped RMSNorm~\cite{wu2018groupnormalization}. Global RMSNorm couples all channels to a single statistic, meaning one outlier suppresses all features. BlockNorm decouples feature subspaces, preventing catastrophic cancellation (which can prove important~\cite{jin2025massivevaluesselfattentionmodules,xiao2024smoothquantaccurateefficientposttraining}) and increasing the model's representational capacity via more independent degrees of freedom. Because block size is strictly consistent between training and inference, the model natively adapts to this grouped topology without requiring global, hardware-inefficient synchronization. Empirically, BlockNorm improves stability over baselines (Figure~\ref{fig:loka_mods_ablation}), and convergence is insensitive to block size provided it is sufficiently large (e.g. 256) and identical across train/test phases. This approach also aligns closely with emerging Microscaling (MX) hardware standards, which utilize block-shared scaling to preserve dynamic range without the overhead of global synchronization~\cite{rouhani2023microscalingdataformatsdeep}.}
The idea is also supported by prior work in the machine learning community. For example, pRMSNorm was proposed alongside RMSNorm~\cite{zhang2019rootmeansquarelayer} which assumes the identical distribution of the neurons and estimates RMS using as little as 6.25\% of them. GroupNorm also found that normalizing over groups of output neurons remains effective for training stability and model quality. \shepherd{To demonstrate this}, \revision{we compare training of a production Wukong model using a BlockNorm of size 256 with RMSNorm in Figure~\ref{fig:blocknorm256}, demonstrate its ability to preserve model quality.}

% However, a potential drawback of this finer-grained normalization is the suppression of massive activation values, which recent LLM studies~\cite{jin2025massivevaluesselfattentionmodules, xiao2024smoothquantaccurateefficientposttraining} have found to be useful. We belive this is acceptable, as we hypothesize that the model can learn to adapt to this suppression, and train/inference mismatch wouldn't be an issue as we keep the same block dim. Because the massive values tend to occur on a few specific channels, the subsequent fully connected layers can learn to re-amplify these channels as needed.

% Moreover, from a low-level numerical precision perspective, BlockNorm can actually improve accuracy by enabling normalization to be performed directly in FP32 precision on intermediate results within the kernel.

\subsubsection{Hard Swish}
Sigmoid-based swish activation and normalization come with large overheads and numerical challenges due to the expensive exponential operations~\cite{fishman2024scaling}, making Swish significantly slower than lightweight activations such as ReLU.

We replace normal swish function with Hard Swish~\cite{pydimarry2024evaluatingmodelperformancehardswish}, a piecewise linear, close approximation to Swish that significantly reduces computational cost while maintaining similar representational power. The Hard Swish function is defined as follows: \[ \text{h-swish}(x) = x \cdot \frac{\code{ReLU6}(x + 3)}{6} \]

where ReLU6 provides a bounded ReLU operation that outputs zero for negative inputs, % the identity for inputs between 0 and 6, and 6 for larger inputs. , as defined below:

%\[
%\text{ReLU6}(x) = \begin{cases} 
%0 & \text{if } x < 0 \\
%x & \text{if } 0 \leq x \leq 6 \\
%6 & \text{if } x > 6
%\end{cases}
%\]

\begin{comment}
\begin{figure}[t]
    \begin{center}
    \includegraphics[width=0.8\linewidth]{Figures/swish_and_hardswish.pdf}
    \caption{Swish and Hard Swish Function}
    \label{fig:swish_and_hard_swish}        
    \end{center}
    \vspace{-1em}
\end{figure}
\end{comment}

The elimination of exponential operations reduces computational overhead, while the piecewise linear nature of Hard Swish makes it naturally amenable to low-precision computation. Additionally, Hard Swish integrates seamlessly with BlockNorm, allowing both operations to be fused within the same kernel for maximum efficiency.

The representational power of Hard Swish remains comparable to standard Swish for the typical input ranges encountered in recommendation models, while providing substantially better behavior under \shepherd{limited numerical range}. This trade-off of slightly simplified activation dynamics in exchange for dramatically improved low-precision stability proves highly beneficial in practice.

\subsection{\shepherd{\lokadispatch: Per-Operator Kernel Orchestration}}

\begin{comment}
    \begin{figure}[t]
   \begin{center}
    \includegraphics[width=1\linewidth]{Figures/dispatch_grid.pdf}
    \caption{Example \lokadispatch Grids}
    \label{fig:dispatch_grid}        
    \end{center}
\end{figure}
\end{comment}

With improved low-precision accuracy enabled by \lokaprobe's targeted analysis and \lokamodules' architectural improvements, the final challenge is achieving optimal end-to-end throughput. Even with numerically stable components, the diversity of GEMM shapes, hardware characteristics, and available libraries means that no single low-precision implementation performs best across all scenarios.

Modern low-precision ecosystems present a complex optimization landscape. Libraries differ fundamentally in their design philosophy: some provide bare kernels with maximum agility (e.g., DeepGEMM), others offer comprehensive quantization pipelines (e.g., FBGEMM), while still others function as drop-in replacements for standard layers (TorchAO). Within each library, multiple recipes provide fine-grained control over forward and backward pass datatypes, quantization strategies (tensorwise, rowwise, blockwise), and other features like fast accumulation.

This diversity, while providing flexibility, creates a challenging optimization problem. Selecting the optimal combination of library and recipe for each GEMM operation requires balancing accuracy constraints with performance objectives, a task that becomes intractable to perform manually across hundreds of modules in production models.

\noindent \textbf{Library and Recipe Selection} \lokadispatch addresses this challenge through a lightweight runtime system that leverages the statistical insights from \lokaprobe to make optimal kernel selection decisions. Rather than applying uniform low-precision policies across all operations, \lokadispatch treats each GEMM as an independent optimization opportunity, selecting the fastest implementation that satisfies accuracy requirements.

The core algorithm operates through constrained optimization. For each GEMM operation, candidate implementations are first filtered by accuracy constraints derived from \lokaprobe's MERE analysis. Only libraries and recipes whose expected error falls below a conservative threshold (typically MERE $< 0.2$) and whose speedup (per \lokaprobe analysis) exceeds a minimum improvement factor (typically $> 1.05\times$) are considered. From this filtered set, \lokadispatch selects the implementation with the highest measured throughput.
The heterogeneous nature of this mapping underscores the importance of fine-grained, per-operation optimization rather than global policies. % Figure~\ref{fig:dispatch_grid} shows an example \lokadispatch grid: given a model, for each batch size and hardware, and for each input and output size, \loka finds the optimal low-precision kernel to use.

\noindent \textbf{Trainer Integration} The technical implementation of \lokadispatch centers on providing a unified interface across diverse low-precision libraries while maintaining compatibility with existing training frameworks. We implement a custom autograd function that serves as a universal adapter, allowing the same high-level interface to route to different underlying kernel implementations.

During model initialization, we perform a transformation pass that identifies target linear layers and replaces them with LoKA-aware linear wrappers. These wrappers encapsulate the dispatch logic while maintaining identical semantics to standard PyTorch linear layers, ensuring compatibility with existing training pipelines and optimization strategies.
The autograd function called by the wrappers handles both forward and backward pass routing, maintaining separate optimization decisions for each direction when beneficial. This granular control proves particularly valuable because forward and backward passes often have different optimal implementations due to varying tensor shapes, layouts, datatypes and computational patterns.

\revision{It's worth noting that \lokadispatch does not need to operate dynamically per-iteration once static profiling results are obtained. For inference, kernel selection is entirely statically determined, and distribution shifts are handled naturally via online continuous training. For training, Dispatch's dynamism acts purely as a safeguard against massive distribution shifts (e.g., sudden user behavior changes during a holiday). Dynamically switching precision mid-training does incur a recompilation tax and is often not worthwhile.}

% The integration of \lokaprobe, \lokamodules, and \lokadispatch creates a comprehensive framework that transforms low-precision computation from a fragile optimization into a reliable performance enhancement. By systematically addressing accuracy analysis, architectural stability, and runtime optimization, LoKA enables production-scale deployment of low-precision training for large recommendation models.

\section{Evaluation}

We comprehensively evaluate \loka across representative LRM architectures in production settings, demonstrating its effectiveness in enabling stable low-precision training and inference while delivering substantial throughput improvements. Our evaluation addresses three key questions: Can \loka achieve lossless low-precision training? What end-to-end performance gains does it deliver across different scales and hardware? How do individual \loka components contribute to overall effectiveness?

\subsection{Setup}
Our evaluation focuses on three families of state-of-the-art LRM architectures that represent different points in the design space: Wukong~\cite{zhang2024wukongscalinglawlargescale}, InterFormer~\cite{interformer} and External Large Foundation Model~\cite{liang2025externallargefoundationmodel}. This selection provides a complete coverage of industry-scale recommenders that also assess open-source components (e.g., Wukong's FMB and LCB~\cite{dcpp}, InterFormer's Transformer~\cite{vaswani2017attention} components, and integration of DCN~\cite{wang2021dcn}, DHEN~\cite{zhang2022dhendeephierarchicalensemble}, DLRM~\cite{dlrm} and SUM~\cite{Zhang_2024} architecture in ELFM). All experiments are conducted on industry-scale datasets that contains tens of billions of examples, each with thousands of features, following the practice in~\cite{zhang2024wukongscalinglawlargescale}.

We conduct experiments across diverse hardware configurations (Nvidia H100/B200 and AMD MI300X clusters with 16-256 GPUs) to capture performance characteristics across different generations and scales. Our implementation integrates LoKA with the PyTorch framework and supports three major low-precision libraries: TorchAO, DeepGEMM, and FBGEMM. All experiments focus on FP8 training and inference, representing the most practical current low-precision target for production deployment. We use TorchRec~\cite{ivchenko2022torchrec}, hybrid parallelism~\cite{neo} with balanced sharding~\cite{neuroshard} and FSDP~\cite{fsdp} for training to minimize communication overheads. We also enable quantized communication~\cite{yang2020training} for all models in BF16 format regardless of compute datatype to ensure fair comparisons across all experiments.

We evaluate \loka across three model configurations and model families representing different scales and computational characteristics in Table~\ref{tab:model_setup}.

\begin{table}[h]
\footnotesize
\centering
\resizebox{1\columnwidth}{!}{%
\begin{tabular}{l c c ccc}
\toprule
Model & \multirow{2}{*}{GFLOPs/sample} & \multirow{2}{*}{Params (B)} & \multicolumn{3}{c}{Batch Size \& GPU Count} \\
Family &  &  & H100 & B200/MI 300X & GB200/MI 350X \\
\midrule
Wukong & 24 & 257  & 6K\&32 & 12K\&16  & 20K\&16 \\
Interformer & 28 & 566  & 4K\&64 & 8K\&64 & 20K\&32 \\
ELFM & 40 & 1343 & 2k\&256 & 6K\&128  & 20K\&32 \\
\bottomrule
\end{tabular}%
}
\vspace{1em}
\caption{Model specifications and Training Setup}
\label{tab:model_setup}
\vspace{-1em}
\end{table}

The experimental methodology emphasizes production realism. We use actual training data (our dataset contains thousands of features), realistic batch sizes, and production-scale model configurations. This approach ensures that our results reflect the performance characteristics that would be observed in real world scenarios. 

\subsection{Lossless FP8 LRM Training}

\begin{figure}[t]
    \begin{center}
    \includegraphics[width=0.9\linewidth]{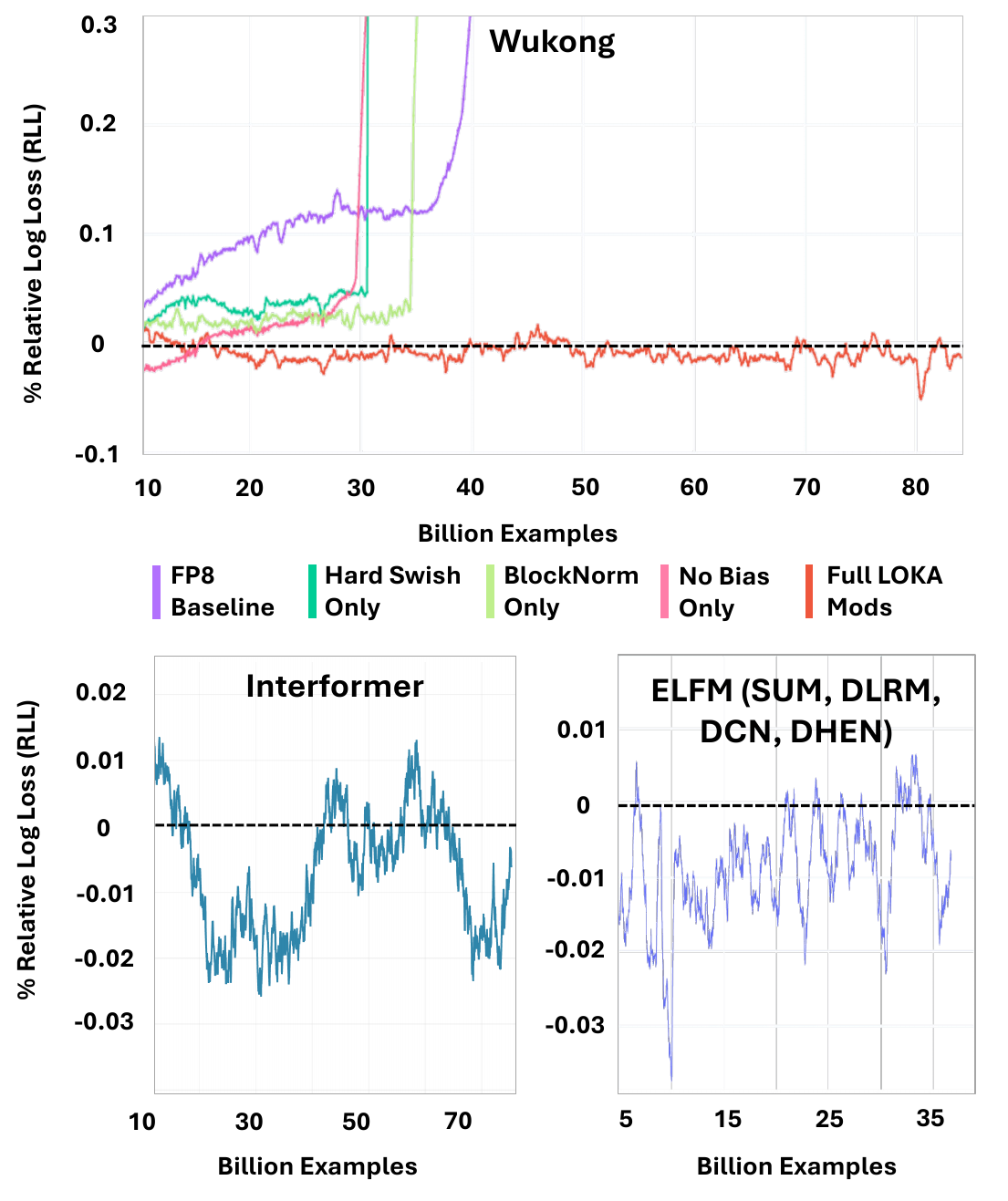}
    \caption{Lossless full-trajectory FP8 training of Wukong, Interformer and ELFM with \loka.}
    \label{fig:ne_parity}       
    \end{center}
    \vspace{-2em}
\end{figure}

The fundamental test of \loka effectiveness is whether it can enable stable FP8 training without accuracy degradation. Figure~\ref{fig:ne_parity} demonstrates this capability on the three models, compared to original, unmodified baselines. We plotted the relative log loss curve with respect to the baseline, excluding model warmup time and we train enough examples until the relative log loss curve converges. Evidently, \loka was not only able to achieve quality neutrality at the end, it was able to do so during the full training trajectory, meaning that within each sampling window, \loka has delivered on par quality consistently. This is especially important for LRMs that process streaming data with shifting distributions. Notably, Figure~\ref{fig:ne_parity} (top) repeated the same model setup, and the dramatic FP8 training failure in Figure~\ref{fig:og_rll} has succeeded with \loka.

\noindent\textbf{Ablation} To quantify the contribution to training stability from each \lokamodule, we quantify its effect in Figure~\ref{fig:ne_parity} (top): no Bias reduces early training instability, BlockNorm provides better numerical conditioning, and Hard Swish reduces activation-related errors. However, none of these modifications alone achieves full stability. Instead, when all \lokamodules are combined, the full system achieves complete loss neutrality throughout training, matching the accuracy of high-precision baselines while operating in FP8. 

\subsection{End-to-end Speedup}

\begin{figure*}[t]
    \centering
    \includegraphics[width=0.85\linewidth]{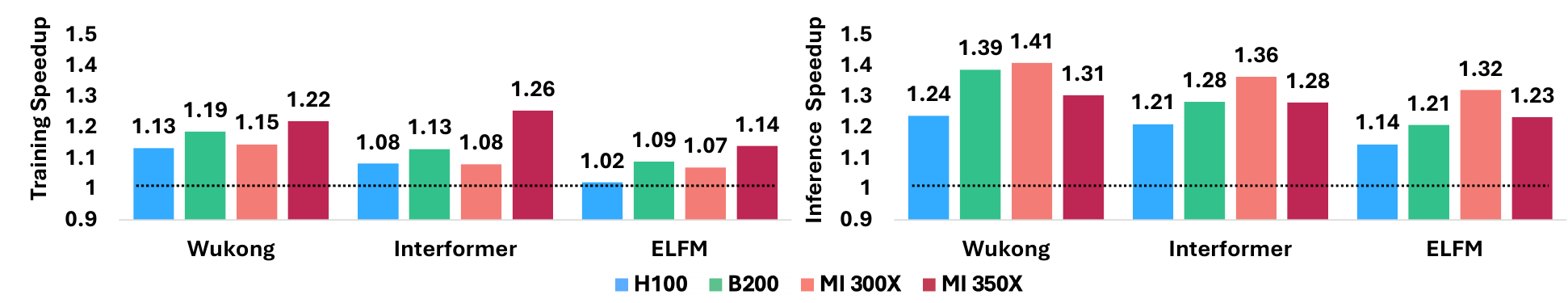}
    \caption{End-to-end Speedup of \loka Training (Left) and Inference (Right)}
    \label{fig:e2e_speedup}
    \vspace{-1em}    
\end{figure*}

Figure~\ref{fig:e2e_speedup} shows the end-to-end training and (in-trainer) inference speedup results, over strong baselines (i.e., models that have been rewritten with \lokamodules), revealing several important patterns. LoKA achieves speedups up to 1.19$\times$ and 1.4$\times$ respectively in training and inference, with performance gains varying systematically based on model and hardware characteristics.

\loka's effectiveness correlates strongly with the compute-to-communication ratio. Models trained at smaller scales with higher computational intensity benefit more from low-precision acceleration, while larger-scale training with more communication overhead sees reduced benefits. 

More recent GPU architectures (B200 vs H100) show larger \loka speedups, primarily due to increased memory capacity enabling larger batch sizes which improves the cost-benefit ratio of low-precision computation.

All tested model architectures benefit substantially from \loka, demonstrating that the framework generalizes across different LRM design approaches rather than being specific to particular architectural choices.

\begin{figure}[t]
    \centering
    \includegraphics[width=0.9\linewidth]{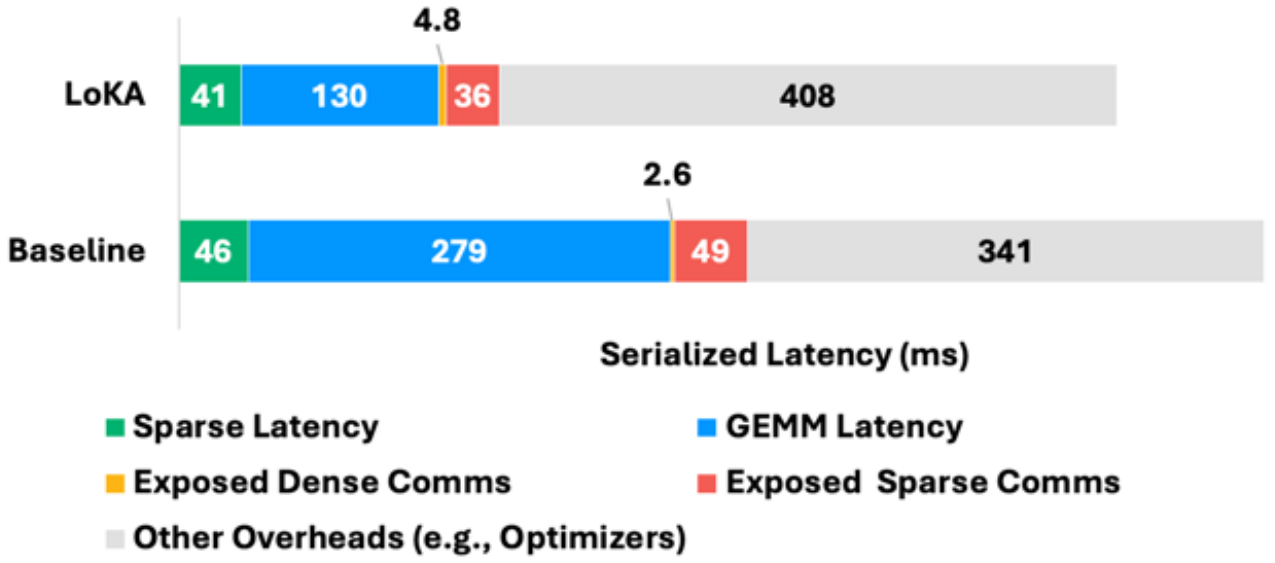}
    \caption{End-to-end latency breakdown of training with and without \loka}
    \label{fig:e2e_breakdown}
\end{figure}

\noindent\textbf{Ablation.}
To quantify LoKA’s contribution to end-to-end training latency, we profile the iteration-level runtime breakdown before and after applying LoKA, using iteration 706 (a representative iteration with stable throughput) averaged across all ranks. Figure~\ref{fig:e2e_breakdown} reports the component-wise latency for the Wukong model trained on 16~B200 GPUs. The majority of the observed speedup originates from reduced GEMM latency (approximately $2\times$). Since LoKA does not modify the sparse embedding pipeline or inter-GPU communication pattern, the slight difference in sparse communication time is attributed to normal run-to-run variability rather than systematic change.

\begin{figure*}[t]
    \centering
    \includegraphics[width=0.85\linewidth]{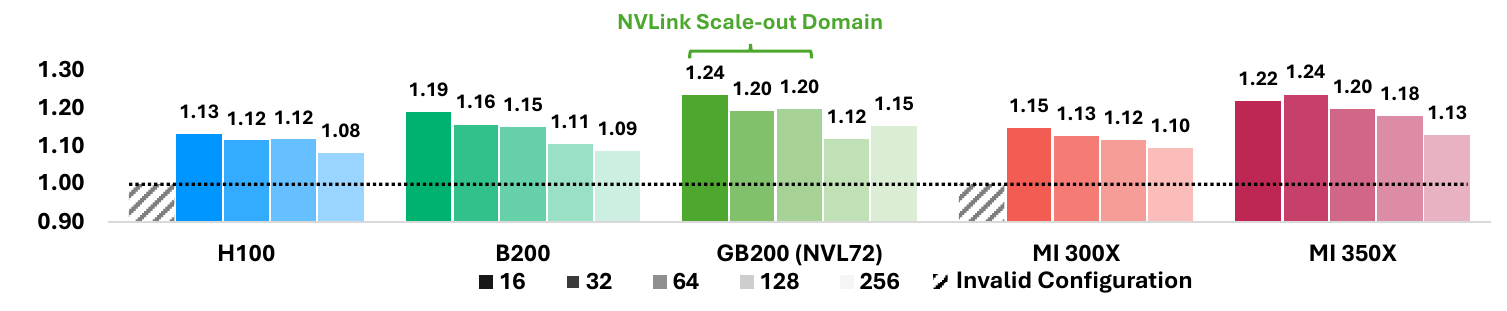}
    \caption{Scalability of \loka on Wukong training, varying number of GPUs. N/A: configuration invalid.}
    \label{fig:e2e_scalability}
    \vspace{-1.5em}
\end{figure*}

\noindent\textbf{Scalability.}
We further evaluate end-to-end throughput scalability using the same Wukong model on 16--256 GPUs. As shown in Figure~\ref{fig:e2e_scalability}, LoKA maintains substantial performance gains at scale, achieving a 10\% throughput improvement at 256~GPUs despite increased communication overhead. While the relative gain diminishes with larger cluster sizes, primarily due to the rising proportion of synchronization and embedding communication~\cite{dmt,luo2020plink,luo2018parameter,278382}, the results confirm LoKA’s effectiveness and practicality in large-scale distributed training.

\subsection{Component Analysis}
We conduct detailed component-level studies examining each \loka component's impact on both accuracy and performance on representative benchmarks.

\subsubsection{\lokaprobe}
We validate \lokaprobe's core value of higher sensitivity in detecting numerical errors through realistic distribution modeling by expanding the MERE measurements study in Table~\ref{tab:rll_bf16_fbgemm_rll} to cover all 27 linear layers from the subject model and summarize the overall results in Table~\ref{tab:lokaprobe_debugger}.

\begin{table}[t]
\footnotesize
\centering
\resizebox{0.8\columnwidth}{!}{%
\begin{tabular}{lcc}
\toprule
 & \textbf{Faulty \revision{Test} Code} & \textbf{Fixed \revision{Test} Code} \\
\midrule
Normal Distribution & 0.42 & 0.42 \\
\lokaprobe & \textbf{17.04} & 0.37 \\
\bottomrule
\end{tabular}%
}
\caption{\revision{Numerical error detection via \lokaprobe's MERE}.}
\label{tab:lokaprobe_debugger}
\vspace{-2em}
\end{table}

Coincidentally, with the help of \lokaprobe, we were also able to discover a faulty implementation in the FBGEMM library's production benchmark: when generating tests with random inputs, the MEREs from correct and incorrect implementation are almost identical, but when using \lokaprobe-generated inputs, the MEREs differed by 47$\times$, which prompted us to investigate and fix \revision{with its developers}.

\subsubsection{\lokamodules}
\begin{figure}[t]
    \begin{center}
    \includegraphics[width=0.9\linewidth]{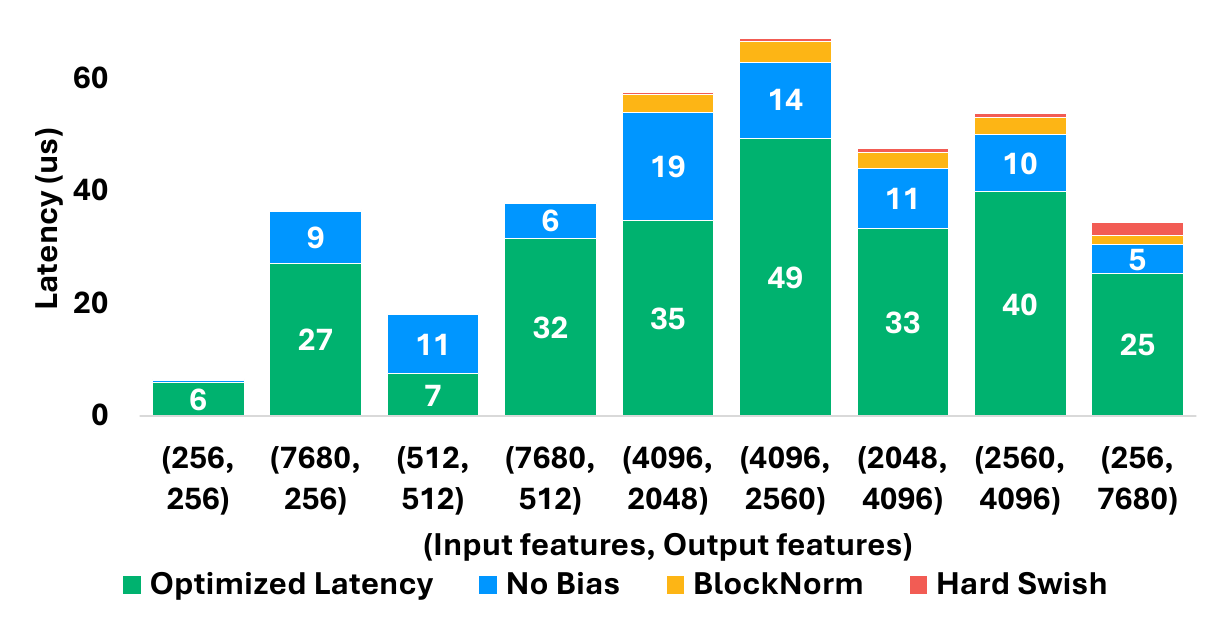}
    \caption{\revision{Assessing \lokamodules' effectiveness on reducing latency of common GEMM sizes, by ablating the latency reduction of each component.}}
    \label{fig:loka_mods_ablation}        
    \end{center}
    \vspace{-2em}    
\end{figure}

Figure~\ref{fig:loka_mods_ablation} quantifies the individual and combined effects of \lokamodules on computational latency using representative GEMM shapes with batch size 1024 (typical \revision{choice}), with torch compile~\cite{torchcompile} enabled. The analysis separates linear layer latency from activation and normalization overhead to isolate the impact of each modification. No Bias provides the largest single contribution to latency reduction, eliminating parameter overhead and simplifying computation paths. BlockNorm delivers substantial improvements by enabling fusion optimizations impossible with standard LayerNorm. Hard Swish contributes meaningful but smaller gains through elimination of expensive exponential operations. The combined effect achieves over 2$\times$ latency reduction.

% with No Bias contributing most significantly, followed by BlockNorm and HardSwish. %This hierarchy reflects both the computational overhead of each operation and the optimization opportunities each modification enables.

% The contribution of each member in \lokamodules to loss parity is highlighted in Figure~\ref{fig:ne_parity}: when used separately, No Bias, BlockNorm and Hard Swish all contributed to training stability under FP8, but none of them is able to achieve lossless FP8 training alone and it requires all of them to present in order to avoid training divergence.

% To understand \lokamodules's effectiveness in latency reduction, we first observe the overall linear latency on a set of typical GEMM sizes, followed by the activation and normalzation block Swish Layernorm, with a batch size of 1024, representing inference settings. We then individually measure the latency achieved by applying No Bias, replacing LayerNorm with BlockNorm and switching to Hard Swish. The result is summarized in Figure~\ref{fig:loka_mods_ablation}: \lokamodules is able to reduce reduce latency by more than 2$\times$ in certain cases, and overall, No Bias contributed to the latency reduction the most, followed by BlockNorm and Hard Swish.

\subsubsection{\lokadispatch}
To isolate \lokadispatch's contribution, we compare it against uniform application of single-library recipes using the same Wukong setup in compute-only mode (eliminating communication effects). The baseline approaches apply consistent recipes across all GEMM operations: TorchAO with tensorwise (TW), rowwise (RW), or a mixed recipe that uses rowwise scaling in the forward but high precision for input gradient and tensorwise scaling for weight in the backward (RW GW HP); DeepGEMM with blockwise (BW) scaling; and FBGEMM with rowwise scaling.

As shown in Table~\ref{tab:loka_dispatch_eval}, \lokadispatch's 1.12$\times$ speedup exceeds the best single-recipe approach (1.08$\times$), demonstrating the value of per-operation optimization. The performance advantage stems from matching each GEMM's characteristics to the most appropriate implementation rather than applying uniform policies that may be suboptimal for specific tensor shapes or computational patterns.

\begin{table}[h]
\footnotesize
\centering
\resizebox{1\columnwidth}{!}{%
\begin{tabular}{lcccccc}
\toprule
 & \multicolumn{3}{c}{TorchAO} & DeepGEMM & FBGEMM & \textbf{\lokadispatch} \\
 & TW & RW & RW GW HP & BW & RW  & Mixed \\
\midrule
 & 1.05 & 1.01 & 1.08 & 0.85 & 0.98 & \textbf{1.12} \\
\bottomrule
\end{tabular}%
}
\caption{\revision{Speedup of \lokadispatch}}
\label{tab:loka_dispatch_eval}
\vspace{-2em}
\end{table}

% The result is summarized in Table~\ref{tab:loka_dispatch_eval}: in the compute-only benchmark, \lokadispatch is able to deliver 1.12$\times$ speedup over the BF16 baseline, significantly higher than the second best's 1.08$\times$ using TorchAO's RW GW HP recipe.

\section{Production Deployments}
We have successfully deployed \loka to \revision{the flagship models of} production advertising recommendation services at a major social media company \revision{serving billions of users}. \revision{\loka has demonstrated 5-20\% end-to-end training throughput and 10-17\% speedup in production inference across various launches}.
% Multiple A/B tests of \loka training are currently underway, showing 5-20\% end-to-end training throughput improvements across various model scales. We are developing more robust tooling and automation for \loka-based workflows to accelerate adoption. % and improve fleet-wide efficiency.

%We have been rolling out deployment of \loka to our Ads recommendation services at a major social media company. We have successfully launched \loka inference to models similar to the ones evaluated, delivering a 17\% inference speedup. 

% Various A/B testing of \loka training are underway, with 5\%-15\% end-to-end training throughput improvements on models across various scales. We are working on more robust tooling support and automation of \loka-based workflow to accelerate \loka adoption and improve our fleet-wide efficiency.
\section{Further Discussions}
\subsection{Related Work}
% \noindent\textbf{Quantization} Numerous quantization approaches have been proposed to improve model training and inference efficiency, ranging from post-training quantization (PTQ)~\cite{banner2019posttraining4bitquantizationconvolution,xiao2024smoothquantaccurateefficientposttraining}, quantization-aware training (QAT)~\cite{jacob2017quantizationtrainingneuralnetworks} to mixed-precision methods~\cite{micikevicius2018mixedprecisiontraining}. Comprehensive surveys~\cite{gholami2021surveyquantizationmethodsefficient, nagel2021whitepaperneuralnetwork} summarize their trade-offs. Recent work extends quantization into low-bit formats such as FP8~\cite{nagel2021whitepaperneuralnetwork} and FP4~\cite{castro2025quartetnativefp4training}, with DeepSeek demonstrating one of the first such applications to LLM training and inference~\cite{deepseekai2025deepseekv3technicalreport}, but these primarily target homogeneous transformer workloads in the LLM context. In contrast, \loka focuses on LRMs, where small-GEMM workloads, normalization overheads, and statistical heterogeneity render existing quantization strategies ineffective.

\noindent\textbf{Quantization.}
Early work on quantized neural networks demonstrated that models can be trained end-to-end with low-precision weights and activations~\cite{hubara2016quantizedneuralnetworkstraining}.
Building on this foundation, a large body of research has explored quantization to improve model efficiency across training and inference.
Existing methods span post-training quantization (PTQ)~\cite{MLSYS2024_42a452cb, banner2019posttraining4bitquantizationconvolution,xiao2024smoothquantaccurateefficientposttraining},
quantization-aware training (QAT)~\cite{jacob2017quantizationtrainingneuralnetworks},
and mixed-precision techniques~\cite{micikevicius2018mixedprecisiontraining},
each balancing accuracy, retraining cost, and implementation complexity.
Comprehensive surveys~\cite{gholami2021surveyquantizationmethodsefficient,nagel2021whitepaperneuralnetwork}
and adaptive schemes such as layer-wise or adaptive quantization~\cite{dumitru2024layerwisequantizationpragmaticeffective,faghri2020adaptivegradientquantizationdataparallel}
further analyze these trade-offs.
More recently, direct low-precision (DLP) training has emerged,
pushing computation into low-bit formats such as FP8 and FP4.
Examples include industrial-scale FP8 deployments~\cite{deepseekai2025deepseekv3technicalreport}
and native FP4 training frameworks~\cite{castro2025quartetnativefp4training,chmiel2025fp4wayfullyquantized,fp4nvda}.
While these efforts establish the feasibility of sub-8-bit arithmetic,
they focus on homogeneous transformer or CNN workloads.
In contrast, \loka targets large-scale recommendation models whose small-GEMM kernels,
normalization-heavy dataflow, and statistical heterogeneity
violate the assumptions underlying existing quantization strategies.

\noindent\textbf{Libraries.}
Several production-grade libraries implement low-precision inference and training across diverse hardware backends.
FBGEMM~\cite{khudia2021fbgemmenablinghighperformancelowprecision} pioneered high-performance
INT8 and FP16 kernels optimized for x86 and GPU architectures, forming the foundation of many industrial recommendation deployments.
DeepGEMM~\cite{deepgemm} and TorchAO~\cite{or2025torchaopytorchnativetrainingtoservingmodel}
extend this direction to FP8 training, providing fine-grained scaling recipes at both CUDA and PyTorch-level.
Other frameworks such as NVIDIA's Transformer Engine~\cite{GitHubNV41:online} and AMD Quark~\cite{Introduc93:online} expose similar quantization abstractions.

While these systems significantly advance kernel-level efficiency,
they rely on synthetic benchmarks and generic GEMM shapes for accuracy validation, which often diverge from real LRM distributions.
As shown in~\cite{gholami2021surveyquantizationmethodsefficient}, kernel accuracy depends strongly on input statistics,
a factor largely unmodeled in current implementations.
Consequently, production LRMs continue to train predominantly in BF16 or TF32 precision,with FP8 adoption limited to post-training inference.
\loka complements these libraries by introducing a statistical probing and runtime selection layer (\lokaprobe) that characterizes per-layer distributions online and dispatches to the optimal low-precision kernel at runtime (\lokadispatch). %, turning these static toolkits into components of a dynamic model-system co-optimization pipeline.

\shepherd{\noindent\textbf{Generalizability.}}
\loka demonstrates strong generalizability across a wide spectrum of large recommendation model architectures and operators.
Its techniques apply seamlessly to the factorization-machine and linear-compression modules in Wukong~\cite{zhang2024wukongscalinglawlargescale},
the transformer-based sequence processors and interleaved non-sequential components in InterFormer~\cite{interformer},
as well as the compound architectures used by ELFM~\cite{liang2025externallargefoundationmodel},
which integrate DHEN~\cite{zhang2022dhendeephierarchicalensemble}, SUM~\cite{Zhang_2024}, DLRM~\cite{naumov2019deeplearningrecommendationmodel}, FMB~\cite{dcpp} and DCN~\cite{wang2021dcn}.
\loka operates directly on the common computational primitives of LRMs—linear transformations, normalization, and activation functions—
making its methods agnostic to architectural topology.
This property allows \loka to extend naturally to emerging hybrid or foundation-scale recommendation models
that combine sequential, graph-based, and multimodal components.

\shepherd{A key test of \loka's principles is whether they transfer to hardware unseen during framework development. \loka was designed and validated on H100, B200, and MI300X clusters. Subsequently, without any modification to \loka's methodology or components, we evaluated on GB200 NVL72 and MI350X — hardware that was not available during \loka's development. As shown in Figures~\ref{fig:e2e_speedup} and ~\ref{fig:e2e_scalability}, \loka delivered comparable speedups on both new platforms, confirming that its principles are not overfit to specific hardware characteristics.}

\subsection{Limitations}
\lokaprobe's benchmarks cannot fully capture real-world performance due to compute-communication overlap, though SM carveout~\cite{smcarveout} techniques help mitigate this. The framework requires models to use standard building blocks, as novel architectures may need additional kernel development. \lokadispatch may require manual intervention when introducing new low-precision kernels to ensure proper integration with compilation frameworks like torch.compile~\cite{torchcompile} for best performance. Additionally, error-based probing~\cite{MLSYS2024_42a452cb} including \loka lacks a mechanism to reason about error propagation throughout the network, making it more conservative than needed: for example, even large errors at each operator may cancel out at the end leading to neutral quality, but \lokadispatch will disable low precision for all such layers, leading to \revision{potential} missed opportunities.

\subsection{Future Work}
We focus on FP8 as the most practical current target for production deployment. While FP4 and other ultra-low precisions show promise, they remain active research areas for dense model training~\cite{fp4nvda,chmiel2025fp4wayfullyquantized,wang2025optimizinglargelanguagemodel,castro2025quartetnativefp4training,hubara2016quantizedneuralnetworkstraining,esser2020learnedstepsizequantization,nvidia2025pretraininglargelanguagemodels}. Recent study shows \revision{Random Hadamard Transform~\cite{le2026scoutattendsketchandwalksparse, ashkboos2024quarot, ashkboos2025halo, deepseekai2025deepseekv32pushingfrontieropen} and quantized communication\cite{Enabling85:online} can further improve \loka efficacy. Orthogonally, AutoML~\cite{wang2019haq,deng2026bitschipsllmbasedhardwareaware} may aid in bridging the gap between MERE and final accuracy.} %that \loka can also aid in understanding which parameters incur low loss under quantization and hence are suitable candidates for quantization.

\section{Conclusion}
Applying low-precision training to LRMs faces challenges from numerical instability and suboptimal throughput gains. \shepherd{We propose \loka, a systematic framework built on three generalizable principles: (1) distribution-aware profiling to identify where low precision is safe, (2) precision-aware model-hardware co-design to expand where it is safe, and (3) per-operator kernel orchestration to maximize the resulting gains}. Our evaluation across three model families and five GPU architectures demonstrates up to 20\% training throughput improvements and 40\% inference acceleration with no quality loss. \loka has been deployed to flagship models at a major social media company, serving billions of users.

% Our evaluation across various model scales demonstrates up to 20\% training throughput improvements and 40\% inference acceleration with no quality loss, turning low-precision from a liability into a reliable performance lever for LRM at scale. \revision{\loka has been deployed at a major social media company, demonstrating its practicality and effectiveness}.

%Applying low-precision training to LRMs faces challenges from numerical instability and suboptimal throughput gains. We propose \loka, a systematic framework that addresses these challenges \revision{in a systematic manner}. % (1) \lokaprobe systematically identifies quantization vulnerabilities using learned input and weight distributions with statistical significance; (2) \lokamodules provide architectural modifications that improve both execution latency and arithmetic fidelity under low-precision conditions; (3) \lokadispatch optimally routes each GEMM to the fastest kernel based on accuracy and performance characteristics.
%
%%%%%%% -- PAPER CONTENT ENDS -- %%%%%%%%

%%%%%%%%% -- BIB STYLE AND FILE -- %%%%%%%%
\bibliographystyle{IEEEtranS}
\bibliography{refs}

@misc{nvda_a100,
author = {{NVIDIA Corporation}},
  title = {NVIDIA A100 GPU Datasheet},
  url = "https://www.nvidia.com/content/dam/en-zz/Solutions/Data-Center/a100/pdf/nvidia-a100-datasheet.pdf",
month = {},
year = {},
  note = "[Online; accessed 2025-08-25]"
}

@misc{nvda_h100,
author = {{NVIDIA Corporation}},
  title = {NVIDIA H100 GPU Datasheet},
  url = "https://resources.nvidia.com/en-us-hopper-architecture/nvidia-tensor-core-gpu-datasheet?ncid=no-ncid",
month = {},
year = {},
  note = "[Online; accessed 2025-08-25]"
}

@misc{nvda_b200,
author = {{NVIDIA Corporation}},
  title = {NVIDIA B200 GPU Datasheet},
  url = "https://nvdam.widen.net/s/wwnsxrhm2w/blackwell-datasheet-3384703",
month = {},
year = {},
  note = "[Online; accessed 2025-08-25]"
}

@misc{zhang2022dhendeephierarchicalensemble,
      title={DHEN: A Deep and Hierarchical Ensemble Network for Large-Scale Click-Through Rate Prediction}, 
      author={Buyun Zhang and Liang Luo and Xi Liu and Jay Li and Zeliang Chen and Weilin Zhang and Xiaohan Wei and Yuchen Hao and Michael Tsang and Wenjun Wang and Yang Liu and Huayu Li and Yasmine Badr and Jongsoo Park and Jiyan Yang and Dheevatsa Mudigere and Ellie Wen},
      year={2022},
      eprint={2203.11014},
      archivePrefix={arXiv},
      primaryClass={cs.IR},
      url={https://arxiv.org/abs/2203.11014}, 
}

@misc{zhang2024wukongscalinglawlargescale,
      title={Wukong: Towards a Scaling Law for Large-Scale Recommendation}, 
      author={Buyun Zhang and Liang Luo and Yuxin Chen and Jade Nie and Xi Liu and Daifeng Guo and Yanli Zhao and Shen Li and Yuchen Hao and Yantao Yao and Guna Lakshminarayanan and Ellie Dingqiao Wen and Jongsoo Park and Maxim Naumov and Wenlin Chen},
      year={2024},
      eprint={2403.02545},
      archivePrefix={arXiv},
      primaryClass={cs.LG},
      url={https://arxiv.org/abs/2403.02545}, 
}

@inproceedings{10.1145/3580305.3599846,
author = {Tang, Jiaxi and Drori, Yoel and Chang, Daryl and Sathiamoorthy, Maheswaran and Gilmer, Justin and Wei, Li and Yi, Xinyang and Hong, Lichan and Chi, Ed H.},
title = {Improving Training Stability for Multitask Ranking Models in Recommender Systems},
year = {2023},
isbn = {9798400701030},
publisher = {Association for Computing Machinery},
address = {New York, NY, USA},
url = {https://doi.org/10.1145/3580305.3599846},
doi = {10.1145/3580305.3599846},
booktitle = {Proceedings of the 29th ACM SIGKDD Conference on Knowledge Discovery and Data Mining},
pages = {4882–4893},
numpages = {12},
location = {Long Beach, CA, USA},
series = {KDD '23}
}

@InProceedings{pmlr-v235-zhai24a,
  title = 	 {Actions Speak Louder than Words: Trillion-Parameter Sequential Transducers for Generative Recommendations},
  author =       {Zhai, Jiaqi and Liao, Lucy and Liu, Xing and Wang, Yueming and Li, Rui and Cao, Xuan and Gao, Leon and Gong, Zhaojie and Gu, Fangda and He, Jiayuan and Lu, Yinghai and Shi, Yu},
  booktitle = 	 {Proceedings of the 41st International Conference on Machine Learning},
  pages = 	 {58484--58509},
  year = 	 {2024},
  editor = 	 {Salakhutdinov, Ruslan and Kolter, Zico and Heller, Katherine and Weller, Adrian and Oliver, Nuria and Scarlett, Jonathan and Berkenkamp, Felix},
  volume = 	 {235},
  series = 	 {Proceedings of Machine Learning Research},
  month = 	 {21--27 Jul},
  publisher =    {PMLR},
  url = 	 {https://proceedings.mlr.press/v235/zhai24a.html}
}

@misc{or2025torchaopytorchnativetrainingtoservingmodel,
      title={TorchAO: PyTorch-Native Training-to-Serving Model Optimization}, 
      author={Andrew Or and Apurva Jain and Daniel Vega-Myhre and Jesse Cai and Charles David Hernandez and Zhenrui Zheng and Driss Guessous and Vasiliy Kuznetsov and Christian Puhrsch and Mark Saroufim and Supriya Rao and Thien Tran and Aleksandar Samardžić},
      year={2025},
      eprint={2507.16099},
      archivePrefix={arXiv},
      primaryClass={cs.LG},
      url={https://arxiv.org/abs/2507.16099}, 
}

@article{neo,
  title={High-performance, distributed training of large-scale deep learning recommendation models},
  author={Mudigere, Dheevatsa and Hao, Yuchen and Huang, Jianyu and Tulloch, Andrew and Sridharan, Srinivas and Liu, Xing and Ozdal, Mustafa and Nie, Jade and Park, Jongsoo and Luo, Liang and others},
  journal={arXiv preprint arXiv:2104.05158},
  year={2021}
}

@article{fsdp,
  title={Pytorch FSDP: experiences on scaling fully sharded data parallel},
  author={Zhao, Yanli and Gu, Andrew and Varma, Rohan and Luo, Liang and Huang, Chien-Chin and Xu, Min and Wright, Less and Shojanazeri, Hamid and Ott, Myle and Shleifer, Sam and others},
  journal={arXiv preprint arXiv:2304.11277},
  year={2023}
}

@article{neuroshard,
  title={Pre-train and Search: Efficient Embedding Table Sharding with Pre-trained Neural Cost Models},
  author={Zha, Daochen and Feng, Louis and Luo, Liang and Bhushanam, Bhargav and Liu, Zirui and Hu, Yusuo and Nie, Jade and Huang, Yuzhen and Tian, Yuandong and Kejariwal, Arun and others},
  journal={Proceedings of Machine Learning and Systems},
  volume={5},
  year={2023}
}

@article{vaswani2017attention,
  title={Attention is all you need},
  author={Vaswani, Ashish and Shazeer, Noam and Parmar, Niki and Uszkoreit, Jakob and Jones, Llion and Gomez, Aidan N and Kaiser, {\L}ukasz and Polosukhin, Illia},
  journal={Advances in neural information processing systems},
  volume={30},
  year={2017}
}

@article{dlrm,
  title={Deep learning recommendation model for personalization and recommendation systems},
  author={Naumov, Maxim and Mudigere, Dheevatsa and Shi, Hao-Jun Michael and Huang, Jianyu and Sundaraman, Narayanan and Park, Jongsoo and Wang, Xiaodong and Gupta, Udit and Wu, Carole-Jean and Azzolini, Alisson G and others},
  journal={arXiv preprint arXiv:1906.00091},
  year={2019}
}

@inproceedings{wang2021dcn,
  title={Dcn v2: Improved deep \& cross network and practical lessons for web-scale learning to rank systems},
  author={Wang, Ruoxi and Shivanna, Rakesh and Cheng, Derek and Jain, Sagar and Lin, Dong and Hong, Lichan and Chi, Ed},
  booktitle={Proceedings of the web conference 2021},
  pages={1785--1797},
  year={2021}
}

@article{dcpp,
  title        = "Dot Product Matrix Compression for Machine Learning",
  author       = {Anonymous},
  journal={Technical Disclosure Commons},
  year={2019},
  howpublished = "\url{https://www.tdcommons.org/cgi/viewcontent.cgi?article=3891&context=dpubs_series}"
}

@article{kaplan2020scaling,
  title={Scaling laws for neural language models},
  author={Kaplan, Jared and McCandlish, Sam and Henighan, Tom and Brown, Tom B and Chess, Benjamin and Child, Rewon and Gray, Scott and Radford, Alec and Wu, Jeffrey and Amodei, Dario},
  journal={arXiv preprint arXiv:2001.08361},
  year={2020}
}

@inproceedings{luo2018parameter,
  title={Parameter hub: a rack-scale parameter server for distributed deep neural network training},
  author={Luo, Liang and Nelson, Jacob and Ceze, Luis and Phanishayee, Amar and Krishnamurthy, Arvind},
  booktitle={Proceedings of the ACM Symposium on Cloud Computing},
  pages={41--54},
  year={2018}
}

@inproceedings{afn,
  title={Adaptive factorization network: Learning adaptive-order feature interactions},
  author={Cheng, Weiyu and Shen, Yanyan and Huang, Linpeng},
  booktitle={Proceedings of the AAAI Conference on Artificial Intelligence},
  volume={34},
  number={04},
  pages={3609--3616},
  year={2020}
}

@article{dmt,
  title={Disaggregated Multi-Tower: Topology-aware Modeling Technique for Efficient Large-Scale Recommendation},
  author={Luo, Liang and Zhang, Buyun and Tsang, Michael and Ma, Yinbin and Chu, Ching-Hsiang and Chen, Yuxin and Li, Shen and Hao, Yuchen and Zhao, Yanli and Lakshminarayanan, Guna and others},
  journal={arXiv preprint arXiv:2403.00877},
  year={2024}
}

@article{ardalani2022understanding,
  title={Understanding scaling laws for recommendation models},
  author={Ardalani, Newsha and Wu, Carole-Jean and Chen, Zeliang and Bhushanam, Bhargav and Aziz, Adnan},
  journal={arXiv preprint arXiv:2208.08489},
  year={2022}
}

@article{guo2023embedding,
  title={On the Embedding Collapse when Scaling up Recommendation Models},
  author={Guo, Xingzhuo and Pan, Junwei and Wang, Ximei and Chen, Baixu and Jiang, Jie and Long, Mingsheng},
  journal={arXiv preprint arXiv:2310.04400},
  year={2023}
}

@article{geng2023vip5,
  title={Vip5: Towards multimodal foundation models for recommendation},
  author={Geng, Shijie and Tan, Juntao and Liu, Shuchang and Fu, Zuohui and Zhang, Yongfeng},
  journal={arXiv preprint arXiv:2305.14302},
  year={2023}
}

@article{luo2020plink,
  title={Plink: Discovering and exploiting locality for accelerated distributed training on the public cloud},
  author={Luo, Liang and West, Peter and Nelson, Jacob and Krishnamurthy, Arvind and Ceze, Luis},
  journal={Proceedings of Machine Learning and Systems},
  volume={2},
  pages={82--97},
  year={2020}
}

@misc{ba2016layernormalization,
      title={Layer Normalization}, 
      author={Jimmy Lei Ba and Jamie Ryan Kiros and Geoffrey E. Hinton},
      year={2016},
      eprint={1607.06450},
      archivePrefix={arXiv},
      primaryClass={stat.ML},
      url={https://arxiv.org/abs/1607.06450}, 
}

@inproceedings{ivchenko2022torchrec,
  title={Torchrec: a pytorch domain library for recommendation systems},
  author={Ivchenko, Dmytro and Van Der Staay, Dennis and Taylor, Colin and Liu, Xing and Feng, Will and Kindi, Rahul and Sudarshan, Anirudh and Sefati, Shahin},
  booktitle={Proceedings of the 16th ACM Conference on Recommender Systems},
  pages={482--483},
  year={2022}
}

@inproceedings{torchcompile,
author = {Ansel, Jason and Yang, Edward and He, Horace and Gimelshein, Natalia and Jain, Animesh and Voznesensky, Michael and Bao, Bin and Bell, Peter and Berard, David and Burovski, Evgeni and Chauhan, Geeta and Chourdia, Anjali and Constable, Will and Desmaison, Alban and DeVito, Zachary and Ellison, Elias and Feng, Will and Gong, Jiong and Gschwind, Michael and Hirsh, Brian and Huang, Sherlock and Kalambarkar, Kshiteej and Kirsch, Laurent and Lazos, Michael and Lezcano, Mario and Liang, Yanbo and Liang, Jason and Lu, Yinghai and Luk, C. K. and Maher, Bert and Pan, Yunjie and Puhrsch, Christian and Reso, Matthias and Saroufim, Mark and Siraichi, Marcos Yukio and Suk, Helen and Zhang, Shunting and Suo, Michael and Tillet, Phil and Zhao, Xu and Wang, Eikan and Zhou, Keren and Zou, Richard and Wang, Xiaodong and Mathews, Ajit and Wen, William and Chanan, Gregory and Wu, Peng and Chintala, Soumith},
title = {PyTorch 2: Faster Machine Learning Through Dynamic Python Bytecode Transformation and Graph Compilation},
year = {2024},
isbn = {9798400703850},
publisher = {Association for Computing Machinery},
address = {New York, NY, USA},
url = {https://doi.org/10.1145/3620665.3640366},
doi = {10.1145/3620665.3640366},
booktitle = {Proceedings of the 29th ACM International Conference on Architectural Support for Programming Languages and Operating Systems, Volume 2},
pages = {929–947},
numpages = {19},
location = {La Jolla, CA, USA},
series = {ASPLOS '24}
}

@misc{interformer,
      title={InterFormer: Towards Effective Heterogeneous Interaction Learning for Click-Through Rate Prediction}, 
      author={Zhichen Zeng and Xiaolong Liu and Mengyue Hang and Xiaoyi Liu and Qinghai Zhou and Chaofei Yang and Yiqun Liu and Yichen Ruan and Laming Chen and Yuxin Chen and Yujia Hao and Jiaqi Xu and Jade Nie and Xi Liu and Buyun Zhang and Wei Wen and Siyang Yuan and Kai Wang and Wen-Yen Chen and Yiping Han and Huayu Li and Chunzhi Yang and Bo Long and Philip S. Yu and Hanghang Tong and Jiyan Yang},
      year={2024},
      eprint={2411.09852},
      archivePrefix={arXiv},
      primaryClass={cs.IR},
      url={https://arxiv.org/abs/2411.09852}, 
}

@article{Ramachandran2017SwishAS,
  title={Swish: a Self-Gated Activation Function},
  author={Prajit Ramachandran and Barret Zoph and Quoc V. Le},
  journal={arXiv: Neural and Evolutionary Computing},
  year={2017},
  url={https://api.semanticscholar.org/CorpusID:196158220}
}

@misc{liang2024torchtitanonestoppytorchnative,
      title={TorchTitan: One-stop PyTorch native solution for production ready LLM pre-training}, 
      author={Wanchao Liang and Tianyu Liu and Less Wright and Will Constable and Andrew Gu and Chien-Chin Huang and Iris Zhang and Wei Feng and Howard Huang and Junjie Wang and Sanket Purandare and Gokul Nadathur and Stratos Idreos},
      year={2024},
      eprint={2410.06511},
      archivePrefix={arXiv},
      primaryClass={cs.CL},
      url={https://arxiv.org/abs/2410.06511}, 
}

@misc{deepseekai2025deepseekv3technicalreport,
      title={DeepSeek-V3 Technical Report}, 
      author={DeepSeek-AI and Aixin Liu and Bei Feng and Bing Xue and Bingxuan Wang and Bochao Wu and Chengda Lu and Chenggang Zhao and Chengqi Deng and Chenyu Zhang and Chong Ruan and Damai Dai and Daya Guo and Dejian Yang and Deli Chen and Dongjie Ji and Erhang Li and Fangyun Lin and Fucong Dai and Fuli Luo and Guangbo Hao and Guanting Chen and Guowei Li and H. Zhang and Han Bao and Hanwei Xu and Haocheng Wang and Haowei Zhang and Honghui Ding and Huajian Xin and Huazuo Gao and Hui Li and Hui Qu and J. L. Cai and Jian Liang and Jianzhong Guo and Jiaqi Ni and Jiashi Li and Jiawei Wang and Jin Chen and Jingchang Chen and Jingyang Yuan and Junjie Qiu and Junlong Li and Junxiao Song and Kai Dong and Kai Hu and Kaige Gao and Kang Guan and Kexin Huang and Kuai Yu and Lean Wang and Lecong Zhang and Lei Xu and Leyi Xia and Liang Zhao and Litong Wang and Liyue Zhang and Meng Li and Miaojun Wang and Mingchuan Zhang and Minghua Zhang and Minghui Tang and Mingming Li and Ning Tian and Panpan Huang and Peiyi Wang and Peng Zhang and Qiancheng Wang and Qihao Zhu and Qinyu Chen and Qiushi Du and R. J. Chen and R. L. Jin and Ruiqi Ge and Ruisong Zhang and Ruizhe Pan and Runji Wang and Runxin Xu and Ruoyu Zhang and Ruyi Chen and S. S. Li and Shanghao Lu and Shangyan Zhou and Shanhuang Chen and Shaoqing Wu and Shengfeng Ye and Shengfeng Ye and Shirong Ma and Shiyu Wang and Shuang Zhou and Shuiping Yu and Shunfeng Zhou and Shuting Pan and T. Wang and Tao Yun and Tian Pei and Tianyu Sun and W. L. Xiao and Wangding Zeng and Wanjia Zhao and Wei An and Wen Liu and Wenfeng Liang and Wenjun Gao and Wenqin Yu and Wentao Zhang and X. Q. Li and Xiangyue Jin and Xianzu Wang and Xiao Bi and Xiaodong Liu and Xiaohan Wang and Xiaojin Shen and Xiaokang Chen and Xiaokang Zhang and Xiaosha Chen and Xiaotao Nie and Xiaowen Sun and Xiaoxiang Wang and Xin Cheng and Xin Liu and Xin Xie and Xingchao Liu and Xingkai Yu and Xinnan Song and Xinxia Shan and Xinyi Zhou and Xinyu Yang and Xinyuan Li and Xuecheng Su and Xuheng Lin and Y. K. Li and Y. Q. Wang and Y. X. Wei and Y. X. Zhu and Yang Zhang and Yanhong Xu and Yanhong Xu and Yanping Huang and Yao Li and Yao Zhao and Yaofeng Sun and Yaohui Li and Yaohui Wang and Yi Yu and Yi Zheng and Yichao Zhang and Yifan Shi and Yiliang Xiong and Ying He and Ying Tang and Yishi Piao and Yisong Wang and Yixuan Tan and Yiyang Ma and Yiyuan Liu and Yongqiang Guo and Yu Wu and Yuan Ou and Yuchen Zhu and Yuduan Wang and Yue Gong and Yuheng Zou and Yujia He and Yukun Zha and Yunfan Xiong and Yunxian Ma and Yuting Yan and Yuxiang Luo and Yuxiang You and Yuxuan Liu and Yuyang Zhou and Z. F. Wu and Z. Z. Ren and Zehui Ren and Zhangli Sha and Zhe Fu and Zhean Xu and Zhen Huang and Zhen Zhang and Zhenda Xie and Zhengyan Zhang and Zhewen Hao and Zhibin Gou and Zhicheng Ma and Zhigang Yan and Zhihong Shao and Zhipeng Xu and Zhiyu Wu and Zhongyu Zhang and Zhuoshu Li and Zihui Gu and Zijia Zhu and Zijun Liu and Zilin Li and Ziwei Xie and Ziyang Song and Ziyi Gao and Zizheng Pan},
      year={2025},
      eprint={2412.19437},
      archivePrefix={arXiv},
      primaryClass={cs.CL},
      url={https://arxiv.org/abs/2412.19437}, 
}

@misc{zhang2019rootmeansquarelayer,
      title={Root Mean Square Layer Normalization}, 
      author={Biao Zhang and Rico Sennrich},
      year={2019},
      eprint={1910.07467},
      archivePrefix={arXiv},
      primaryClass={cs.LG},
      url={https://arxiv.org/abs/1910.07467}, 
}

@inproceedings{yang2020training,
  title={Training deep learning recommendation model with quantized collective communications},
  author={Yang, Jie Amy and Park, Jongsoo and Sridharan, Srinivas and Tang, Ping Tak Peter},
  booktitle={Conference on Knowledge Discovery and Data Mining (KDD)},
  pages={95},
  year={2020}
}

@misc{gholami2021surveyquantizationmethodsefficient,
      title={A Survey of Quantization Methods for Efficient Neural Network Inference}, 
      author={Amir Gholami and Sehoon Kim and Zhen Dong and Zhewei Yao and Michael W. Mahoney and Kurt Keutzer},
      year={2021},
      eprint={2103.13630},
      archivePrefix={arXiv},
      primaryClass={cs.CV},
      url={https://arxiv.org/abs/2103.13630}, 
}

@misc{khudia2021fbgemmenablinghighperformancelowprecision,
      title={FBGEMM: Enabling High-Performance Low-Precision Deep Learning Inference}, 
      author={Daya Khudia and Jianyu Huang and Protonu Basu and Summer Deng and Haixin Liu and Jongsoo Park and Mikhail Smelyanskiy},
      year={2021},
      eprint={2101.05615},
      archivePrefix={arXiv},
      primaryClass={cs.LG},
      url={https://arxiv.org/abs/2101.05615}, 
}

@misc{deepgemm,
author = {DeepSeek},
  title = {DeepGEMM: clean and efficient FP8 GEMM kernels with fine-grained scaling},
  url = "https://github.com/deepseek-ai/DeepGEMM",
month = {},
year = {},
  note = "[Online; accessed 2025-08-29]"
}

@misc{fp4nvda,
author = {Kirthi Devleker},
  title = {NVFP4 Trains with Precision of 16-Bit and Speed and Efficiency of 4-Bit | NVIDIA Technical Blog},
  url = "https://developer.nvidia.com/blog/nvfp4-trains-with-precision-of-16-bit-and-speed-and-efficiency-of-4-bit/",
month = {8},
year = {2025},
  note = "[Online; accessed 2025-08-31]"
}

@misc{chmiel2025fp4wayfullyquantized,
      title={FP4 All the Way: Fully Quantized Training of LLMs}, 
      author={Brian Chmiel and Maxim Fishman and Ron Banner and Daniel Soudry},
      year={2025},
      eprint={2505.19115},
      archivePrefix={arXiv},
      primaryClass={cs.LG},
      url={https://arxiv.org/abs/2505.19115}, 
}

@misc{wang2025optimizinglargelanguagemodel,
      title={Optimizing Large Language Model Training Using FP4 Quantization}, 
      author={Ruizhe Wang and Yeyun Gong and Xiao Liu and Guoshuai Zhao and Ziyue Yang and Baining Guo and Zhengjun Zha and Peng Cheng},
      year={2025},
      eprint={2501.17116},
      archivePrefix={arXiv},
      primaryClass={cs.LG},
      url={https://arxiv.org/abs/2501.17116}, 
}

@article{chan1983algorithms,
  title={Algorithms for computing the sample variance: Analysis and recommendations},
  author={Chan, Tony F and Golub, Gene H and LeVeque, Randall J},
  journal={The American Statistician},
  volume={37},
  number={3},
  pages={242--247},
  year={1983},
  publisher={Taylor \& Francis}
}

@book{gupta2018matrix,
  title={Matrix variate distributions},
  author={Gupta, Arjun K and Nagar, Daya K},
  year={2018},
  publisher={Chapman and Hall/CRC}
}

@misc{harville1998matrix,
  title={Matrix algebra from a statistician's perspective},
  author={Harville, David A},
  year={1998},
  publisher={Taylor \& Francis}
}

@misc{anil2023palm2technicalreport,
      title={PaLM 2 Technical Report}, 
      author={Rohan Anil and Andrew M. Dai and Orhan Firat and Melvin Johnson and Dmitry Lepikhin and Alexandre Passos and Siamak Shakeri and Emanuel Taropa and Paige Bailey and Zhifeng Chen and Eric Chu and Jonathan H. Clark and Laurent El Shafey and Yanping Huang and Kathy Meier-Hellstern and Gaurav Mishra and Erica Moreira and Mark Omernick and Kevin Robinson and Sebastian Ruder and Yi Tay and Kefan Xiao and Yuanzhong Xu and Yujing Zhang and Gustavo Hernandez Abrego and Junwhan Ahn and Jacob Austin and Paul Barham and Jan Botha and James Bradbury and Siddhartha Brahma and Kevin Brooks and Michele Catasta and Yong Cheng and Colin Cherry and Christopher A. Choquette-Choo and Aakanksha Chowdhery and Clément Crepy and Shachi Dave and Mostafa Dehghani and Sunipa Dev and Jacob Devlin and Mark Díaz and Nan Du and Ethan Dyer and Vlad Feinberg and Fangxiaoyu Feng and Vlad Fienber and Markus Freitag and Xavier Garcia and Sebastian Gehrmann and Lucas Gonzalez and Guy Gur-Ari and Steven Hand and Hadi Hashemi and Le Hou and Joshua Howland and Andrea Hu and Jeffrey Hui and Jeremy Hurwitz and Michael Isard and Abe Ittycheriah and Matthew Jagielski and Wenhao Jia and Kathleen Kenealy and Maxim Krikun and Sneha Kudugunta and Chang Lan and Katherine Lee and Benjamin Lee and Eric Li and Music Li and Wei Li and YaGuang Li and Jian Li and Hyeontaek Lim and Hanzhao Lin and Zhongtao Liu and Frederick Liu and Marcello Maggioni and Aroma Mahendru and Joshua Maynez and Vedant Misra and Maysam Moussalem and Zachary Nado and John Nham and Eric Ni and Andrew Nystrom and Alicia Parrish and Marie Pellat and Martin Polacek and Alex Polozov and Reiner Pope and Siyuan Qiao and Emily Reif and Bryan Richter and Parker Riley and Alex Castro Ros and Aurko Roy and Brennan Saeta and Rajkumar Samuel and Renee Shelby and Ambrose Slone and Daniel Smilkov and David R. So and Daniel Sohn and Simon Tokumine and Dasha Valter and Vijay Vasudevan and Kiran Vodrahalli and Xuezhi Wang and Pidong Wang and Zirui Wang and Tao Wang and John Wieting and Yuhuai Wu and Kelvin Xu and Yunhan Xu and Linting Xue and Pengcheng Yin and Jiahui Yu and Qiao Zhang and Steven Zheng and Ce Zheng and Weikang Zhou and Denny Zhou and Slav Petrov and Yonghui Wu},
      year={2023},
      eprint={2305.10403},
      archivePrefix={arXiv},
      primaryClass={cs.CL},
      url={https://arxiv.org/abs/2305.10403}, 
}

@misc{almazrouei2023falconseriesopenlanguage,
      title={The Falcon Series of Open Language Models}, 
      author={Ebtesam Almazrouei and Hamza Alobeidli and Abdulaziz Alshamsi and Alessandro Cappelli and Ruxandra Cojocaru and Mérouane Debbah and Étienne Goffinet and Daniel Hesslow and Julien Launay and Quentin Malartic and Daniele Mazzotta and Badreddine Noune and Baptiste Pannier and Guilherme Penedo},
      year={2023},
      eprint={2311.16867},
      archivePrefix={arXiv},
      primaryClass={cs.CL},
      url={https://arxiv.org/abs/2311.16867}, 
}

@misc{smwaveeffect,
author = {},
  title = {Matrix Multiplication Background User's Guide - NVIDIA Docs},
  url = "https://docs.nvidia.com/deeplearning/performance/dl-performance-matrix-multiplication/index.html#wave-quant",
month = {},
year = {},
  note = "[Online; accessed 2025-09-11]"
}

@misc{wu2018groupnormalization,
      title={Group Normalization}, 
      author={Yuxin Wu and Kaiming He},
      year={2018},
      eprint={1803.08494},
      archivePrefix={arXiv},
      primaryClass={cs.CV},
      url={https://arxiv.org/abs/1803.08494}, 
}

@misc{jin2025massivevaluesselfattentionmodules,
      title={Massive Values in Self-Attention Modules are the Key to Contextual Knowledge Understanding}, 
      author={Mingyu Jin and Kai Mei and Wujiang Xu and Mingjie Sun and Ruixiang Tang and Mengnan Du and Zirui Liu and Yongfeng Zhang},
      year={2025},
      eprint={2502.01563},
      archivePrefix={arXiv},
      primaryClass={cs.CL},
      url={https://arxiv.org/abs/2502.01563}, 
}

@misc{pydimarry2024evaluatingmodelperformancehardswish,
      title={Evaluating Model Performance with Hard-Swish Activation Function Adjustments}, 
      author={Sai Abhinav Pydimarry and Shekhar Madhav Khairnar and Sofia Garces Palacios and Ganesh Sankaranarayanan and Darian Hoagland and Dmitry Nepomnayshy and Huu Phong Nguyen},
      year={2024},
      eprint={2410.06879},
      archivePrefix={arXiv},
      primaryClass={cs.CV},
      url={https://arxiv.org/abs/2410.06879}, 
}

@misc{castro2025quartetnativefp4training,
      title={Quartet: Native FP4 Training Can Be Optimal for Large Language Models}, 
      author={Roberto L. Castro and Andrei Panferov and Soroush Tabesh and Oliver Sieberling and Jiale Chen and Mahdi Nikdan and Saleh Ashkboos and Dan Alistarh},
      year={2025},
      eprint={2505.14669},
      archivePrefix={arXiv},
      primaryClass={cs.LG},
      url={https://arxiv.org/abs/2505.14669}, 
}

@misc{smcarveout,
author = {Nvidia},
  title = {1. NVIDIA Ampere GPU Architecture Tuning Guide — Ampere Tuning Guide 13.0 documentation},
  url = "https://docs.nvidia.com/cuda/ampere-tuning-guide/index.html",
month = {},
year = {},
  note = "[Online; accessed 2025-09-12]"
}

@article{gao2025deck,
  title={DECK: Experiences on Delta Checkpointing for Industrial Recommendation Systems},
  author={Gao, Xin and Acharya, Sibasish and Han, Sihui and Ren, Yongxiong and Zhao, Yanli and Luo, Liang and Wang, Chucheng and Fernando, Pradeep and Mishra, Saurabh and Yan, Siqi and others},
  journal={Proceedings of the VLDB Endowment},
  volume={18},
  number={12},
  pages={4978--4990},
  year={2025},
  publisher={VLDB Endowment}
}

@misc{liang2025externallargefoundationmodel,
      title={External Large Foundation Model: How to Efficiently Serve Trillions of Parameters for Online Ads Recommendation}, 
      author={Mingfu Liang and Xi Liu and Rong Jin and Boyang Liu and Qiuling Suo and Qinghai Zhou and Song Zhou and Laming Chen and Hua Zheng and Zhiyuan Li and Shali Jiang and Jiyan Yang and Xiaozhen Xia and Fan Yang and Yasmine Badr and Ellie Wen and Shuyu Xu and Hansey Chen and Zhengyu Zhang and Jade Nie and Chunzhi Yang and Zhichen Zeng and Weilin Zhang and Xingliang Huang and Qianru Li and Shiquan Wang and Evelyn Lyu and Wenjing Lu and Rui Zhang and Wenjun Wang and Jason Rudy and Mengyue Hang and Kai Wang and Yinbin Ma and Shuaiwen Wang and Sihan Zeng and Tongyi Tang and Xiaohan Wei and Longhao Jin and Jamey Zhang and Marcus Chen and Jiayi Xu and Angie Huang and Xihuan Zeng and Chi Zhang and Zhengli Zhao and Jared Yang and Qiang Jin and Xian Chen and Amit Anand Amlesahwaram and Lexi Song and Liang Luo and Yuchen Hao and Nan Xiao and Yavuz Yetim and Luoshang Pan and Gaoxiang Liu and Yuxi Hu and Yuzhen Huang and Jackie Xu and Rich Zhu and Xin Zhang and Yiqun Liu and Hang Yin and Yuxin Chen and Buyun Zhang and Xiaoyi Liu and Xingyuan Wang and Wenguang Mao and Zhijing Li and Zhehui Zhou and Feifan Gu and Qin Huang and Chonglin Sun and Nancy Yu and Shuo Gu and Shupin Mao and Benjamin Au and Jingzheng Qin and Peggy Yao and Jae-Woo Choi and Bin Gao and Ernest Wang and Lei Zhang and Wen-Yen Chen and Ted Lee and Yujie Zha and Yi Meng and Alex Gong and Edison Gao and Jack Hsueh and Jie Zheng and Alireza Vahdatpour and Yiping Han and Yantao Yao and Toshinari Kureha and Shuo Chang and Musharaf Sultan and John Bocharov and Sagar Chordia and Xiaorui Gan and Peng Sun and Rocky Liu and Bo Long and Wenlin Chen and Santanu Kolay and Huayu Li},
      year={2025},
      eprint={2502.17494},
      archivePrefix={arXiv},
      primaryClass={cs.IR},
      url={https://arxiv.org/abs/2502.17494}, 
}

@misc{banner2019posttraining4bitquantizationconvolution,
      title={Post-training 4-bit quantization of convolution networks for rapid-deployment}, 
      author={Ron Banner and Yury Nahshan and Elad Hoffer and Daniel Soudry},
      year={2019},
      eprint={1810.05723},
      archivePrefix={arXiv},
      primaryClass={cs.CV},
      url={https://arxiv.org/abs/1810.05723}, 
}

@misc{jacob2017quantizationtrainingneuralnetworks,
      title={Quantization and Training of Neural Networks for Efficient Integer-Arithmetic-Only Inference}, 
      author={Benoit Jacob and Skirmantas Kligys and Bo Chen and Menglong Zhu and Matthew Tang and Andrew Howard and Hartwig Adam and Dmitry Kalenichenko},
      year={2017},
      eprint={1712.05877},
      archivePrefix={arXiv},
      primaryClass={cs.LG},
      url={https://arxiv.org/abs/1712.05877}, 
}

@misc{micikevicius2018mixedprecisiontraining,
      title={Mixed Precision Training}, 
      author={Paulius Micikevicius and Sharan Narang and Jonah Alben and Gregory Diamos and Erich Elsen and David Garcia and Boris Ginsburg and Michael Houston and Oleksii Kuchaiev and Ganesh Venkatesh and Hao Wu},
      year={2018},
      eprint={1710.03740},
      archivePrefix={arXiv},
      primaryClass={cs.AI},
      url={https://arxiv.org/abs/1710.03740}, 
}

@misc{nagel2021whitepaperneuralnetwork,
      title={A White Paper on Neural Network Quantization}, 
      author={Markus Nagel and Marios Fournarakis and Rana Ali Amjad and Yelysei Bondarenko and Mart van Baalen and Tijmen Blankevoort},
      year={2021},
      eprint={2106.08295},
      archivePrefix={arXiv},
      primaryClass={cs.LG},
      url={https://arxiv.org/abs/2106.08295}, 
}

@misc{hubara2016quantizedneuralnetworkstraining,
      title={Quantized Neural Networks: Training Neural Networks with Low Precision Weights and Activations}, 
      author={Itay Hubara and Matthieu Courbariaux and Daniel Soudry and Ran El-Yaniv and Yoshua Bengio},
      year={2016},
      eprint={1609.07061},
      archivePrefix={arXiv},
      primaryClass={cs.NE},
      url={https://arxiv.org/abs/1609.07061}, 
}

@misc{xiao2024smoothquantaccurateefficientposttraining,
      title={SmoothQuant: Accurate and Efficient Post-Training Quantization for Large Language Models}, 
      author={Guangxuan Xiao and Ji Lin and Mickael Seznec and Hao Wu and Julien Demouth and Song Han},
      year={2024},
      eprint={2211.10438},
      archivePrefix={arXiv},
      primaryClass={cs.CL},
      url={https://arxiv.org/abs/2211.10438}, 
}

@misc{dumitru2024layerwisequantizationpragmaticeffective,
      title={Layer-Wise Quantization: A Pragmatic and Effective Method for Quantizing LLMs Beyond Integer Bit-Levels}, 
      author={Razvan-Gabriel Dumitru and Vikas Yadav and Rishabh Maheshwary and Paul-Ioan Clotan and Sathwik Tejaswi Madhusudhan and Mihai Surdeanu},
      year={2024},
      eprint={2406.17415},
      archivePrefix={arXiv},
      primaryClass={cs.CL},
      url={https://arxiv.org/abs/2406.17415}, 
}

@misc{faghri2020adaptivegradientquantizationdataparallel,
      title={Adaptive Gradient Quantization for Data-Parallel SGD}, 
      author={Fartash Faghri and Iman Tabrizian and Ilia Markov and Dan Alistarh and Daniel Roy and Ali Ramezani-Kebrya},
      year={2020},
      eprint={2010.12460},
      archivePrefix={arXiv},
      primaryClass={cs.LG},
      url={https://arxiv.org/abs/2010.12460}, 
}

@misc{esser2020learnedstepsizequantization,
      title={Learned Step Size Quantization}, 
      author={Steven K. Esser and Jeffrey L. McKinstry and Deepika Bablani and Rathinakumar Appuswamy and Dharmendra S. Modha},
      year={2020},
      eprint={1902.08153},
      archivePrefix={arXiv},
      primaryClass={cs.LG},
      url={https://arxiv.org/abs/1902.08153}, 
}

@misc{GitHubNV41:online,
author = {
        NVIDIA
    },
  title = {GitHub - NVIDIA/TransformerEngine: A library for accelerating Transformer models on NVIDIA GPUs, including using 8-bit floating point (FP8) precision on Hopper, Ada and Blackwell GPUs, to provide better performance with lower memory utilization in both training and inference.},
  url = "https://github.com/NVIDIA/TransformerEngine",
month = {},
year = {},
  note = "[Online; accessed 2025-10-29]"
}

@misc{Introduc93:online,
author = {AMD},
  title = {Introduction — AMD Quark 0.10 documentation},
  url = "https://quark.docs.amd.com/latest/onnx/tutorial_microexponents_quantization.html",
month = {},
year = {},
  note = "[Online; accessed 2025-10-29]"
}

@inproceedings{Zhang_2024, series={WWW ’24},
   title={Scaling User Modeling: Large-scale Online User Representations for Ads Personalization in Meta},
   url={http://dx.doi.org/10.1145/3589335.3648301},
   DOI={10.1145/3589335.3648301},
   booktitle={Companion Proceedings of the ACM Web Conference 2024},
   publisher={ACM},
   author={Zhang, Wei and Li, Dai and Liang, Chen and Zhou, Fang and Zhang, Zhongke and Wang, Xuewei and Li, Ru and Zhou, Yi and Huang, Yaning and Liang, Dong and Wang, Kai and Wang, Zhangyuan and Chen, Zhengxing and Wu, Fenggang and Chen, Minghai and Li, Huayu and Wu, Yunnan and Shu, Zhan and Yuan, Mindi and Reddy, Sri},
   year={2024},
   month=may, pages={47–55},
   collection={WWW ’24} }

@misc{naumov2019deeplearningrecommendationmodel,
      title={Deep Learning Recommendation Model for Personalization and Recommendation Systems}, 
      author={Maxim Naumov and Dheevatsa Mudigere and Hao-Jun Michael Shi and Jianyu Huang and Narayanan Sundaraman and Jongsoo Park and Xiaodong Wang and Udit Gupta and Carole-Jean Wu and Alisson G. Azzolini and Dmytro Dzhulgakov and Andrey Mallevich and Ilia Cherniavskii and Yinghai Lu and Raghuraman Krishnamoorthi and Ansha Yu and Volodymyr Kondratenko and Stephanie Pereira and Xianjie Chen and Wenlin Chen and Vijay Rao and Bill Jia and Liang Xiong and Misha Smelyanskiy},
      year={2019},
      eprint={1906.00091},
      archivePrefix={arXiv},
      primaryClass={cs.IR},
      url={https://arxiv.org/abs/1906.00091}, 
}

@misc{Enabling85:online,
author = {Will Feng},
  title = {Enabling Float8 All-Gather in FSDP2 - distributed - PyTorch Developer Mailing List},
  url = "https://dev-discuss.pytorch.org/t/enabling-float8-all-gather-in-fsdp2/2359",
month = {8},
year = {2024},
  note = "[Online; accessed 2025-10-29]"
}

@inproceedings {278382,
author = {Jingrong Chen and Hong Zhang and Wei Zhang and Liang Luo and Jeffrey Chase and Ion Stoica and Danyang Zhuo},
title = {{NetHint}: {White-Box} Networking for {Multi-Tenant} Data Centers},
booktitle = {19th USENIX Symposium on Networked Systems Design and Implementation (NSDI 22)},
year = {2022},
isbn = {978-1-939133-27-4},
address = {Renton, WA},
pages = {1327--1343},
url = {https://www.usenix.org/conference/nsdi22/presentation/chen-jingrong},
publisher = {USENIX Association},
month = apr
}

@inproceedings{MLSYS2024_42a452cb,
 author = {Lin, Ji and Tang, Jiaming and Tang, Haotian and Yang, Shang and Chen, Wei-Ming and Wang, Wei-Chen and Xiao, Guangxuan and Dang, Xingyu and Gan, Chuang and Han, Song},
 booktitle = {Proceedings of Machine Learning and Systems},
 editor = {P. Gibbons and G. Pekhimenko and C. De Sa},
 pages = {87--100},
 title = {AWQ: Activation-aware Weight Quantization for On-Device LLM Compression and Acceleration},
 url = {https://proceedings.mlsys.org/paper_files/paper/2024/file/42a452cbafa9dd64e9ba4aa95cc1ef21-Paper-Conference.pdf},
 volume = {6},
 year = {2024}
}

@misc{deepseekai2025deepseekv32pushingfrontieropen,
      title={DeepSeek-V3.2: Pushing the Frontier of Open Large Language Models}, 
      author={DeepSeek-AI and Aixin Liu and Aoxue Mei and Bangcai Lin and Bing Xue and Bingxuan Wang and Bingzheng Xu and Bochao Wu and Bowei Zhang and Chaofan Lin and Chen Dong and Chengda Lu and Chenggang Zhao and Chengqi Deng and Chenhao Xu and Chong Ruan and Damai Dai and Daya Guo and Dejian Yang and Deli Chen and Erhang Li and Fangqi Zhou and Fangyun Lin and Fucong Dai and Guangbo Hao and Guanting Chen and Guowei Li and H. Zhang and Hanwei Xu and Hao Li and Haofen Liang and Haoran Wei and Haowei Zhang and Haowen Luo and Haozhe Ji and Honghui Ding and Hongxuan Tang and Huanqi Cao and Huazuo Gao and Hui Qu and Hui Zeng and Jialiang Huang and Jiashi Li and Jiaxin Xu and Jiewen Hu and Jingchang Chen and Jingting Xiang and Jingyang Yuan and Jingyuan Cheng and Jinhua Zhu and Jun Ran and Junguang Jiang and Junjie Qiu and Junlong Li and Junxiao Song and Kai Dong and Kaige Gao and Kang Guan and Kexin Huang and Kexing Zhou and Kezhao Huang and Kuai Yu and Lean Wang and Lecong Zhang and Lei Wang and Liang Zhao and Liangsheng Yin and Lihua Guo and Lingxiao Luo and Linwang Ma and Litong Wang and Liyue Zhang and M. S. Di and M. Y Xu and Mingchuan Zhang and Minghua Zhang and Minghui Tang and Mingxu Zhou and Panpan Huang and Peixin Cong and Peiyi Wang and Qiancheng Wang and Qihao Zhu and Qingyang Li and Qinyu Chen and Qiushi Du and Ruiling Xu and Ruiqi Ge and Ruisong Zhang and Ruizhe Pan and Runji Wang and Runqiu Yin and Runxin Xu and Ruomeng Shen and Ruoyu Zhang and S. H. Liu and Shanghao Lu and Shangyan Zhou and Shanhuang Chen and Shaofei Cai and Shaoyuan Chen and Shengding Hu and Shengyu Liu and Shiqiang Hu and Shirong Ma and Shiyu Wang and Shuiping Yu and Shunfeng Zhou and Shuting Pan and Songyang Zhou and Tao Ni and Tao Yun and Tian Pei and Tian Ye and Tianyuan Yue and Wangding Zeng and Wen Liu and Wenfeng Liang and Wenjie Pang and Wenjing Luo and Wenjun Gao and Wentao Zhang and Xi Gao and Xiangwen Wang and Xiao Bi and Xiaodong Liu and Xiaohan Wang and Xiaokang Chen and Xiaokang Zhang and Xiaotao Nie and Xin Cheng and Xin Liu and Xin Xie and Xingchao Liu and Xingkai Yu and Xingyou Li and Xinyu Yang and Xinyuan Li and Xu Chen and Xuecheng Su and Xuehai Pan and Xuheng Lin and Xuwei Fu and Y. Q. Wang and Yang Zhang and Yanhong Xu and Yanru Ma and Yao Li and Yao Li and Yao Zhao and Yaofeng Sun and Yaohui Wang and Yi Qian and Yi Yu and Yichao Zhang and Yifan Ding and Yifan Shi and Yiliang Xiong and Ying He and Ying Zhou and Yinmin Zhong and Yishi Piao and Yisong Wang and Yixiao Chen and Yixuan Tan and Yixuan Wei and Yiyang Ma and Yiyuan Liu and Yonglun Yang and Yongqiang Guo and Yongtong Wu and Yu Wu and Yuan Cheng and Yuan Ou and Yuanfan Xu and Yuduan Wang and Yue Gong and Yuhan Wu and Yuheng Zou and Yukun Li and Yunfan Xiong and Yuxiang Luo and Yuxiang You and Yuxuan Liu and Yuyang Zhou and Z. F. Wu and Z. Z. Ren and Zehua Zhao and Zehui Ren and Zhangli Sha and Zhe Fu and Zhean Xu and Zhenda Xie and Zhengyan Zhang and Zhewen Hao and Zhibin Gou and Zhicheng Ma and Zhigang Yan and Zhihong Shao and Zhixian Huang and Zhiyu Wu and Zhuoshu Li and Zhuping Zhang and Zian Xu and Zihao Wang and Zihui Gu and Zijia Zhu and Zilin Li and Zipeng Zhang and Ziwei Xie and Ziyi Gao and Zizheng Pan and Zongqing Yao and Bei Feng and Hui Li and J. L. Cai and Jiaqi Ni and Lei Xu and Meng Li and Ning Tian and R. J. Chen and R. L. Jin and S. S. Li and Shuang Zhou and Tianyu Sun and X. Q. Li and Xiangyue Jin and Xiaojin Shen and Xiaosha Chen and Xinnan Song and Xinyi Zhou and Y. X. Zhu and Yanping Huang and Yaohui Li and Yi Zheng and Yuchen Zhu and Yunxian Ma and Zhen Huang and Zhipeng Xu and Zhongyu Zhang and Dongjie Ji and Jian Liang and Jianzhong Guo and Jin Chen and Leyi Xia and Miaojun Wang and Mingming Li and Peng Zhang and Ruyi Chen and Shangmian Sun and Shaoqing Wu and Shengfeng Ye and T. Wang and W. L. Xiao and Wei An and Xianzu Wang and Xiaowen Sun and Xiaoxiang Wang and Ying Tang and Yukun Zha and Zekai Zhang and Zhe Ju and Zhen Zhang and Zihua Qu},
      year={2025},
      eprint={2512.02556},
      archivePrefix={arXiv},
      primaryClass={cs.CL},
      url={https://arxiv.org/abs/2512.02556}, 
}

@misc{githubDeepGEMMteststest_bf16pyMain,
	author = {DeepSeek-AI},
	title = {DeepGEMM Numerical Test},
	howpublished = {\url{https://github.com/deepseek-ai/DeepGEMM/blob/main/tests/test_bf16.py#L38}},
	year = {},
	note = {[Accessed 17-02-2026]},
}

@misc{githubAotestkerneltest_blockwise_tritonpyMain,
	author = {TorchAO},
	title = {TorchAO Blockwise Triton Test},
	howpublished = {\url{https://github.com/pytorch/ao/blob/main/test/kernel/test\_blockwise\_triton.py\#L55}},
	year = {},
	note = {[Accessed 17-02-2026]},
}

@misc{githubFBGEMMfbgemm_gputestquantizefused_8bit_rowwise_testpyMain,
	author = {FBGEMM},
	title = {FBGEMM Numerical Test},
	howpublished = {\url{https://github.com/pytorch/FBGEMM/blob/main/fbgemm_gpu/test/quantize/fused_8bit_rowwise_test.py#L61)—rely}},
	year = {},
	note = {[Accessed 17-02-2026]},
}

@inproceedings{10.1145/3315508.3329973,
author = {Tillet, Philippe and Kung, H. T. and Cox, David},
title = {Triton: an intermediate language and compiler for tiled neural network computations},
year = {2019},
isbn = {9781450367196},
publisher = {Association for Computing Machinery},
address = {New York, NY, USA},
url = {https://doi.org/10.1145/3315508.3329973},
doi = {10.1145/3315508.3329973},
booktitle = {Proceedings of the 3rd ACM SIGPLAN International Workshop on Machine Learning and Programming Languages},
pages = {10–19},
numpages = {10},
location = {Phoenix, AZ, USA},
series = {MAPL 2019}
}

@inproceedings{wang2019haq,
  title={Haq: Hardware-aware automated quantization with mixed precision},
  author={Wang, Kuan and Liu, Zhijian and Lin, Yujun and Lin, Ji and Han, Song},
  booktitle={Proceedings of the IEEE/CVF conference on computer vision and pattern recognition},
  pages={8612--8620},
  year={2019}
}

@misc{dettmers2022llmint88bitmatrixmultiplication,
      title={LLM.int8(): 8-bit Matrix Multiplication for Transformers at Scale}, 
      author={Tim Dettmers and Mike Lewis and Younes Belkada and Luke Zettlemoyer},
      year={2022},
      eprint={2208.07339},
      archivePrefix={arXiv},
      primaryClass={cs.LG},
      url={https://arxiv.org/abs/2208.07339}, 
}

@misc{rouhani2023microscalingdataformatsdeep,
      title={Microscaling Data Formats for Deep Learning}, 
      author={Bita Darvish Rouhani and Ritchie Zhao and Ankit More and Mathew Hall and Alireza Khodamoradi and Summer Deng and Dhruv Choudhary and Marius Cornea and Eric Dellinger and Kristof Denolf and Stosic Dusan and Venmugil Elango and Maximilian Golub and Alexander Heinecke and Phil James-Roxby and Dharmesh Jani and Gaurav Kolhe and Martin Langhammer and Ada Li and Levi Melnick and Maral Mesmakhosroshahi and Andres Rodriguez and Michael Schulte and Rasoul Shafipour and Lei Shao and Michael Siu and Pradeep Dubey and Paulius Micikevicius and Maxim Naumov and Colin Verrilli and Ralph Wittig and Doug Burger and Eric Chung},
      year={2023},
      eprint={2310.10537},
      archivePrefix={arXiv},
      primaryClass={cs.LG},
      url={https://arxiv.org/abs/2310.10537}, 
}

@article{fishman2024scaling,
  title={Scaling fp8 training to trillion-token llms},
  author={Fishman, Maxim and Chmiel, Brian and Banner, Ron and Soudry, Daniel},
  journal={arXiv preprint arXiv:2409.12517},
  year={2024}
}

@article{hernandez2025towards,
  title={Towards fully fp8 gemm llm training at scale},
  author={Hern{\'a}ndez-Cano, Alejandro and Garbaya, Dhia and Schlag, Imanol and Jaggi, Martin},
  journal={arXiv preprint arXiv:2505.20524},
  year={2025}
}

@misc{deng2026bitschipsllmbasedhardwareaware,
      title={From Bits to Chips: An LLM-based Hardware-Aware Quantization Agent for Streamlined Deployment of LLMs}, 
      author={Kaiyuan Deng and Hangyu Zheng and Minghai Qing and Kunxiong Zhu and Gen Li and Yang Xiao and Lan Emily Zhang and Linke Guo and Bo Hui and Yanzhi Wang and Geng Yuan and Gagan Agrawal and Wei Niu and Xiaolong Ma},
      year={2026},
      eprint={2601.03484},
      archivePrefix={arXiv},
      primaryClass={cs.LG},
      url={https://arxiv.org/abs/2601.03484}, 
}

@misc{nvidia2025pretraininglargelanguagemodels,
      title={Pretraining Large Language Models with NVFP4}, 
      author={NVIDIA and Felix Abecassis and Anjulie Agrusa and Dong Ahn and Jonah Alben and Stefania Alborghetti and Michael Andersch and Sivakumar Arayandi and Alexis Bjorlin and Aaron Blakeman and Evan Briones and Ian Buck and Bryan Catanzaro and Jinhang Choi and Mike Chrzanowski and Eric Chung and Victor Cui and Steve Dai and Bita Darvish Rouhani and Carlo del Mundo and Deena Donia and Burc Eryilmaz and Henry Estela and Abhinav Goel and Oleg Goncharov and Yugi Guvvala and Robert Hesse and Russell Hewett and Herbert Hum and Ujval Kapasi and Brucek Khailany and Mikail Khona and Nick Knight and Alex Kondratenko and Ronny Krashinsky and Ben Lanir and Simon Layton and Michael Lightstone and Daniel Lo and Paulius Micikevicius and Asit Mishra and Tim Moon and Deepak Narayanan and Chao Ni and Abhijit Paithankar and Satish Pasumarthi and Ankit Patel and Mostofa Patwary and Ashwin Poojary and Gargi Prasad and Sweta Priyadarshi and Yigong Qin and Xiaowei Ren and Oleg Rybakov and Charbel Sakr and Sanjeev Satheesh and Stas Sergienko and Pasha Shamis and Kirthi Shankar and Nishant Sharma and Mohammad Shoeybi and Michael Siu and Misha Smelyanskiy and Darko Stosic and Dusan Stosic and Bor-Yiing Su and Frank Sun and Nima Tajbakhsh and Shelby Thomas and Przemek Tredak and Evgeny Tsykunov and Gandhi Vaithilingam and Aditya Vavre and Rangharajan Venkatesan and Roger Waleffe and Qiyu Wan and Hexin Wang and Mengdi Wang and Lizzie Wei and Hao Wu and Evan Wu and Keith Wyss and Ning Xu and Jinze Xue and Charlene Yang and Yujia Zhai and Ruoxi Zhang and Jingyang Zhu and Zhongbo Zhu},
      year={2025},
      eprint={2509.25149},
      archivePrefix={arXiv},
      primaryClass={cs.CL},
      url={https://arxiv.org/abs/2509.25149}, 
}

@misc{le2026scoutattendsketchandwalksparse,
      title={Scout Before You Attend: Sketch-and-Walk Sparse Attention for Efficient LLM Inference}, 
      author={Hoang Anh Duy Le and Sahil Joshi and Zeyu Yang and Zhaozhuo Xu and Anshumali Shrivastava},
      year={2026},
      eprint={2602.07397},
      archivePrefix={arXiv},
      primaryClass={cs.LG},
      url={https://arxiv.org/abs/2602.07397}, 
}

@article{ashkboos2024quarot,
  title={Quarot: Outlier-free 4-bit inference in rotated llms},
  author={Ashkboos, Saleh and Mohtashami, Amirkeivan and Croci, Maximilian L and Li, Bo and Cameron, Pashmina and Jaggi, Martin and Alistarh, Dan and Hoefler, Torsten and Hensman, James},
  journal={Advances in Neural Information Processing Systems},
  volume={37},
  pages={100213--100240},
  year={2024}
}

@article{ashkboos2025halo,
  title={Halo: Hadamard-assisted lower-precision optimization for llms},
  author={Ashkboos, Saleh and Nikdan, Mahdi and Tabesh, Soroush and Castro, Roberto L and Hoefler, Torsten and Alistarh, Dan},
  journal={arXiv preprint arXiv:2501.02625},
  year={2025}
}
%%%%%%%%%%%%%%%%%%%%%%%%%%%%%%%%%%%%

\end{document}